%% file: main.tex
\documentclass[lettersize,journal]{IEEEtran}
\usepackage{amsmath,amsfonts}
\usepackage{algorithmic}
\usepackage{algorithm}
\usepackage{array}
\usepackage[caption=false,font=normalsize,labelfont=sf,textfont=sf]{subfig}
\usepackage{textcomp}
\usepackage{stfloats}
\usepackage{url}
\usepackage{verbatim}
\usepackage{graphicx}
\usepackage{cite}
\usepackage{tcolorbox}
\usepackage{balance}
\usepackage{booktabs}
\usepackage{arydshln}

\usepackage{pifont}
\usepackage{wrapfig}
\usepackage{svg} 
\usepackage{colortbl}  
\usepackage{amssymb}

\usepackage{multirow}
\usepackage{indentfirst}
\usepackage{threeparttable}
\usepackage{makecell}
\usepackage{tabularx}
\usepackage[table]{xcolor}
\newcommand{\blue}[1]{\textbf{\textcolor{mblue}{#1}}}
\newcommand{\red}[1]{\textbf{\textcolor{mred}{#1}}}
\definecolor{mblue}{RGB}{0, 77, 128}

\definecolor{mblue}{RGB}{0, 77, 128}
\definecolor{mred}{RGB}{192,0, 0}
\definecolor{darkgreen}{rgb}{0.0, 0.5, 0.0}
\definecolor{mycolor_blue}{HTML}{E7EFFA}
\definecolor{mycolor_gray}{HTML}{ECECEC}
\newcommand{\cmark}{\textcolor{darkgreen}{\ding{51}}}  
\newcommand{\xmark}{\textcolor{red}{\ding{55}}}        

\begin{document}

\title{3DGS-VBench+ and 3DGSV-Assessor: A Multi-Dimensional Subjective Benchmark and a Reference-Flexible Objective Model for 3D Gaussian Splatting Video Quality Assessment}

\author{Yuke Xing, William Gordon, Jiarui Wang, Yujie Zhang, Guangtao Zhai,~\IEEEmembership{Fellow,~IEEE,} and Yiling Xu,~\IEEEmembership{Member,~IEEE}
\thanks{Yuke Xing, Jiarui Wang, Guangtao Zhai and Yiling Xu are with the School of Information Science and Electronic Engineering,
Shanghai Jiao Tong University, Shanghai 200240, China (e-mail: {xingyuke-v, wangjiarui, zhaiguangtao, yl.xu}@sjtu.edu.cn).}
\thanks{Yujie Zhang is with JD.COM, Beijing, 100176, China (zhangyujie.1998@jd.com).}
\thanks{William Gordon is with the Basis Independent Silicon Valley, San Jose, CA 95134, USA (e-mail:williamg.research@gmail.com).}
\thanks{Corresponding author: Yiling Xu.}}

\maketitle

\input{sec/0_abstract_opti}
\input{sec/1_introduction_opti_2}
\input{sec/2_related_work}

\input{sec/3_database}

\input{sec/4_method-6}

\input{sec/5_exp_opti}

\input{sec/6_conclusion}

\section*{Acknowledgments}
This paper is supported in part by National Natural Science Foundation of China (62371290), National Key R\&D Program of China (2024YFB2907204), STCSM under Grant(24511107100), Shanghai Key Lab of Digital Media Processing and Transmission, Shanghai Jiao Tong University.

\bibliographystyle{IEEEtran}
\bibliography{main.bib}

\section{Biography Section}
\begin{IEEEbiography}[{\includegraphics[width=1in,height=1.25in,clip,keepaspectratio]{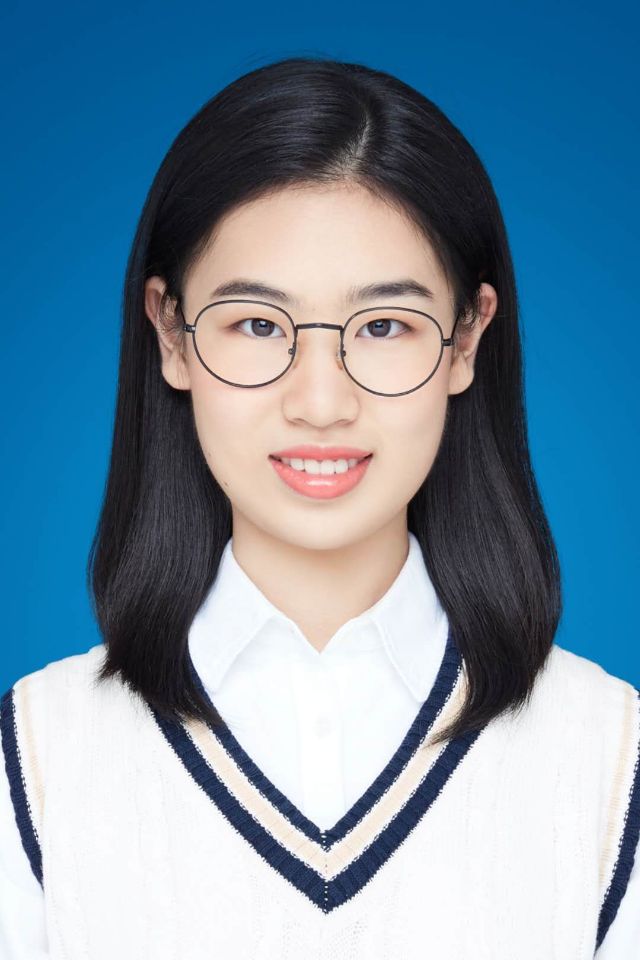}}]{Yuke Xing}
received the B.E. degree in electronic engineering from Shanghai Jiao Tong University, Shanghai, China, in 2024. She is currently working toward the Ph.D. degree with the Cooperative MediaNet Innovation Center, Shanghai Jiao Tong University, Shanghai, China. Her research interests include perceptual quality assessment, 3D novel view synthesis, computer vision, and multimodal signal processing.
\end{IEEEbiography}

\begin{IEEEbiography}[{\includegraphics[width=1in,height=1.25in,clip,keepaspectratio]{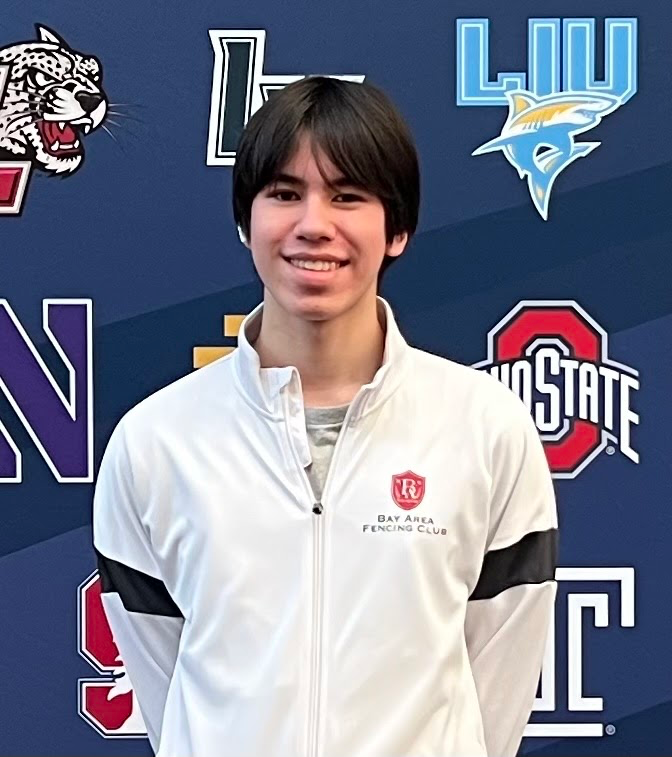}}]{William Gordon}
William Gordon is a junior at Basis Independent Silicon Valley. His research interests include 3D modeling, computer vision,  and quality assessment. His other interests include fencing (with US national raking and medal) and building robotics.
\end{IEEEbiography}

\begin{IEEEbiography}[{\includegraphics[width=1in,height=1.25in,clip,keepaspectratio]{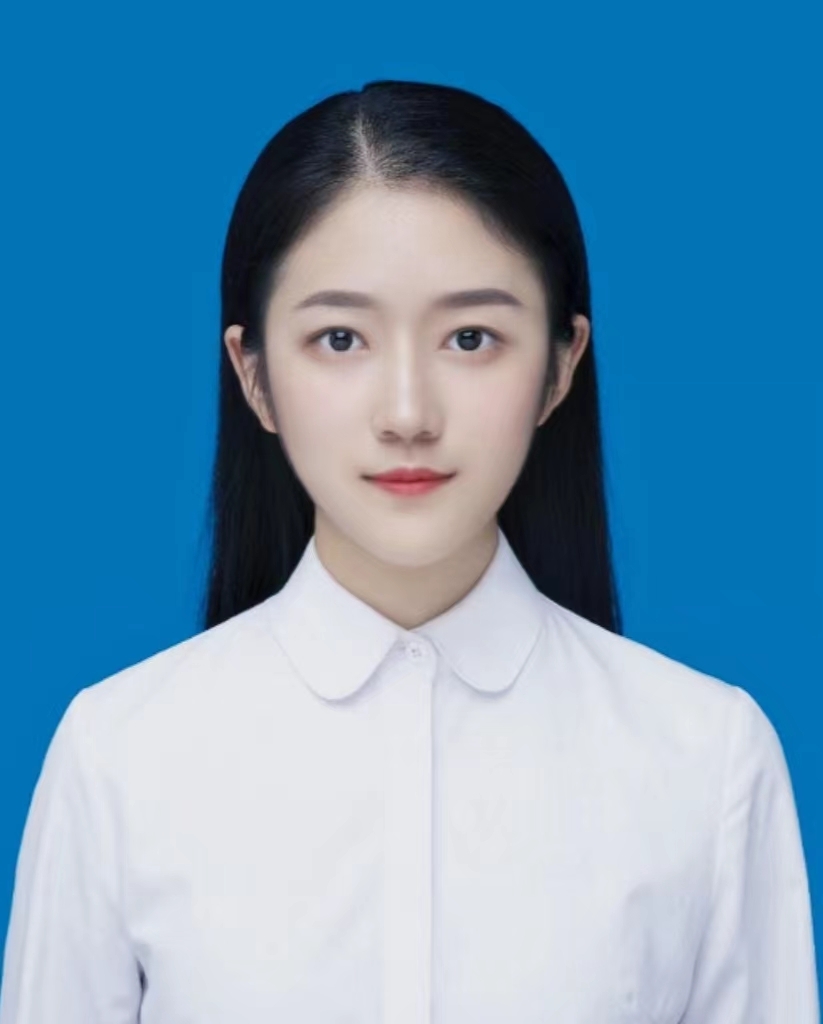}}]{Jiarui Wang}
received the B.E. degree in communication engineering from Shanghai Jiao Tong University, Shanghai, China, in 2024. She is currently working toward the Ph.D. degree with the Department of Electronic Engineering, Shanghai Jiao Tong University, Shanghai, China. Her research interests include perceptual quality assessment, quality of experience, and multimodal signal processing.
\end{IEEEbiography}

\begin{IEEEbiography}[{\includegraphics[width=1in,height=1.25in,clip,keepaspectratio]{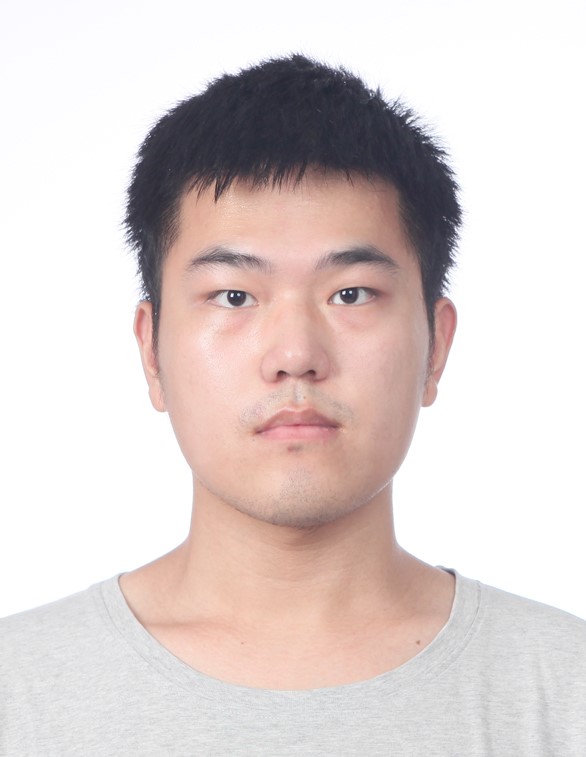}}]{Yujie Zhang}
received the B.S. degree in electronic and information engineering from Beihang University, Beijing, China, in 2020, and the Ph.D. degree in information and communication engineering from Shanghai Jiao Tong University, Shanghai, China, in 2026. He is currently a researcher with JD.com. His research interests include visual quality assessment and reinforcement learning.
\end{IEEEbiography}

\begin{IEEEbiography}[{\includegraphics[width=1in,height=1.25in,clip,keepaspectratio]{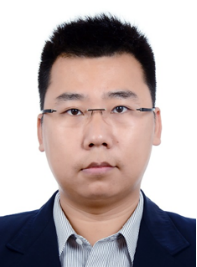}}]{Guangtao Zhai}
 (SM’19) received the B.E. and M.E. degrees from Shandong University, Shandong, China, in 2001 and 2004, respectively, and the Ph.D. degree from Shanghai Jiao Tong University, Shanghai, China, in 2009, where he is currently a Research Professor with the Institute of Image Communication and Information Processing. From 2008 to 2009, he was a Visiting Student with the Department of Electrical and Computer Engineering, McMaster University, Hamilton, ON, Canada, where he was a Post-Doctoral Fellow from 2010 to 2012. From 2012 to 2013, he was a Humboldt Research Fellow with the Institute of Multimedia Communication and Signal Processing, Friedrich Alexander University of Erlangen-Nuremberg, Germany. He received the Award of National Excellent Ph.D. Thesis from the Ministry of Education of China in 2012. His research interests include multimedia signal processing and perceptual signal processing.
\end{IEEEbiography}

\begin{IEEEbiography}[{\includegraphics[width=1in,height=1.25in,clip,keepaspectratio]{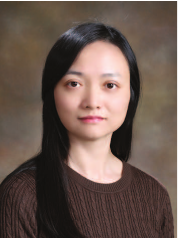}}]{Yiling Xu}
received the B.S., M.S., and Ph.D degrees from the University of Electronic Science
and Technology of China, in 1999, 2001, and 2004
respectively. From 2004 to 2013, she was a senior
engineer with the Multimedia Communication Research Institute, Samsung Electronics Inc., South
Korea. She joined Shanghai Jiao Tong University,
where she is currently a professor in the areas of
multimedia communication, 3D point cloud compression and assessment, system design, and network
optimization. She is the associate editor of the IEEE
Transactions on Broadcasting. She is also an active member in standard
organizations, including MPEG, 3GPP, and AVS.
\end{IEEEbiography}

\vfill

\end{document}

%% file: sec/0_abstract_opti.tex
\begin{abstract} 
3D Gaussian Splatting (3DGS) enables real-time novel view synthesis (NVS), but its heavy storage makes compression essential, and both training and compression introduce representation-specific distortions.
Since generative compression in many methods treats geometric and color attributes via different parameters, it can induce dimension-specific distortions, motivating multi-dimensional assessment.
Moreover, real-world captures generally lack an aligned ground-truth video for a synthesized, continuous camera trajectory, leaving only sparse captured views as references.
To enable multi-dimensional quality assessment of 3DGS videos with scarce aligned references, we first present 3DGS-VBench+, a large-scale subjective video quality assessment (VQA) benchmark comprising 3{,}036 3DGS reconstructions, each paired with a rendered video, across 11 real-world and 22 synthetic scenes, generated by 6 representative compression algorithms at systematically designed levels and annotated with 9{,}108 MOSs covering overall, geometry, and color quality.
Based on 3DGS-VBench+, we propose 3DGSV-Assessor, a reference-flexible, large multimodal framework that scores all three dimensions in one generation pass and fuses a ranking-based reference cue into the frozen regressor through a bounded gate, supporting both reduced-reference (RR) and no-reference (NR) assessment. 
Extensive experiments demonstrate state-of-the-art correlation with human perception on 3DGS-VBench+ and strong zero-shot generalization to other NVS benchmarks. The dataset and code will be released at \url{https://github.com/YukeXing/3DGSV-Assessor}.
\end{abstract}

\begin{IEEEkeywords}
3D Gaussian Splatting, video quality assessment, benchmark, compression, large language model

\end{IEEEkeywords}

%% file: sec/1_introduction_opti_2.tex
\section{Introduction}

\IEEEPARstart{N}{ovel} View Synthesis (NVS) seeks to reconstruct complete scenes from a limited set of captured viewpoints and render images from arbitrary novel perspectives. The field has progressed rapidly from neural radiance fields (NeRF)~\cite{nerf} to 3D Gaussian Splatting (3DGS)~\cite{3DGS}, which replaces implicit volume rendering with an explicit set of anisotropic Gaussians, thereby achieving real-time, high-fidelity rendering~\cite{3DGSsurvey}. However, this explicit form comes at the price of heavy storage footprints that limit practical use, motivating a wave of state-of-the-art (SOTA) 3DGS variants~\cite{Scaffold-GS,HAC,LightGaussian,CompGS,C3dgs,Compact-3DGS} that embed compression directly into the reconstruction pipeline. Despite this significant progress, 3DGS is still far from visually flawless: its training and compression introduce characteristic distortions~\cite{NVS-SQA,ENeRF-QA,3DGS-IEval-15K}. Since human evaluation is expensive and inefficient, it is important to develop reliable 3DGS video quality assessment (VQA) metrics that align well with human perception, both to judge the perceptual quality of a reconstructed scene and to guide downstream applications.

\begin{figure}[t]
    \centering
    \vspace{-1mm}
    \includegraphics[width=1\linewidth]{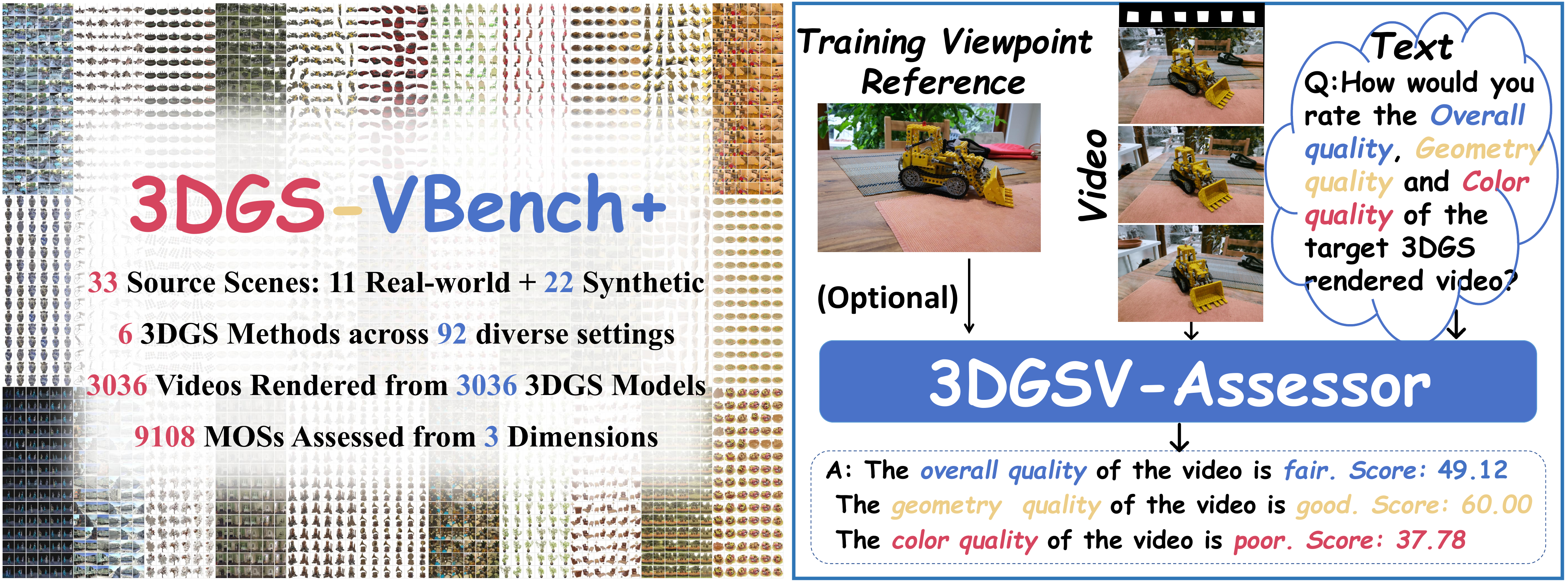}
    \vspace{-5.5mm}
    \caption{Overview of the proposed benchmark and model. Left: 3DGS-VBench+, a multi-dimensional subjective VQA benchmark for compressed 3DGS. Right: 3DGSV-Assessor scores the overall, geometry, and color quality of a rendered video, exploiting a training-viewpoint reference when available and operating no-reference otherwise.}
    \label{figure:teasor1}
    \vspace{-7mm}
\end{figure}

Traditional metrics such as PSNR and SSIM~\cite{PSNR} remain widely used for NVS evaluation, despite their limited correlation with perceived 3DGS quality~\cite{NVS-SQA,GSC-QA,ENeRF-QA,3DGS-IEval-15K}. Developing a VQA metric tailored to 3DGS requires addressing three key challenges. First, the unique representation of 3DGS introduces characteristic artifacts, such as structural omissions, floating and fogging~\cite{3DGS-IEval-15K,ENeRF-QA}. Some of these, such as flickering caused by uneven fitting across training and novel views, further manifest as temporal instability that only a continuous viewing path can expose~\cite{NVS-SQA,ENeRF-QA,sparse-view-3dgs}.
Second, 3DGS compression can affect geometric and color fidelity differently, making an overall quality score alone insufficient. Many 3DGS compression methods separately control the compression of geometric attributes (\textit{e.g.}, scale and rotation) and color attributes (\textit{e.g.}, spherical harmonic (SH) coefficients), resulting in different degrees of geometric and color degradation. These differences motivate separate geometry and color assessments alongside an overall quality judgment, as illustrated in Fig.~\ref{figure:teasor1}.
Third, reference information for 3DGS VQA is often limited and varies across practical scenarios. Scenes are reconstructed from discrete captured views but may be rendered along arbitrary, user-chosen trajectories, for which aligned ground-truth videos are generally unavailable. Full-reference (FR) assessment is therefore often impractical, whereas no-reference (NR) assessment is the common case, and reduced-reference (RR) assessment is feasible because a captured training view can accompany the model as a weak reference at marginal cost. Together, these challenges call for a single, multi-dimensional, \emph{reference-flexible} VQA model that captures 3DGS-specific distortions, exploits the weak reference when available, and remains robust in its absence.

Developing and validating such a model requires a suitable subjective quality benchmark. However, current NVS VQA benchmarks exhibit three key limitations, as summarized in Table~\ref{tab:relate}: \textbf{(1) Insufficient scale.} Most existing datasets contain fewer than 500 videos, largely due to high reconstruction training costs, limiting their support for learning perceptually robust and generalizable models. \textbf{(2) Limited distortion coverage.} Few benchmarks systematically cover different compression strategies and parameter levels, resulting in limited diversity of compression-induced distortions. \textbf{(3) Coarse annotation granularity.} Existing benchmarks provide only overall quality scores, leaving geometric and color quality assessment without dedicated subjective supervision.

To address these gaps, we first present \textbf{3DGS-VBench+}, a large-scale, multi-dimensional subjective VQA benchmark for compressed 3DGS. It contains 3{,}036 3DGS models reconstructed from 11 real-world and 22 synthetic scenes taken from mainstream multi-view datasets~\cite{Mip-NeRF-360,Tanks-Temples,Deep-Blending,ENeRF-QA}. Six representative compression algorithms~\cite{Scaffold-GS,HAC,LightGaussian,CompGS,C3dgs,Compact-3DGS} are applied at systematically designed levels, spanning a broad and controlled range of distortion types and strengths. Each model is rendered into a video along an orbital camera trajectory around the scene. Through a subjective study, we collect mean opinion scores (MOSs) for the overall, geometry, and color dimensions, giving 9{,}108 MOSs in total, and we verify the reliability of the dataset and the diversity of its content.

Building on 3DGS-VBench+, we propose \textbf{3DGSV-Assessor}, a large multimodal model (LMM)-based framework for all-in-one, multi-dimensional 3DGS VQA that supports both NR and RR assessment through three stages. 
Stage~1 builds the foundation: an NR regressor that adapts a vision-language backbone to score all three quality dimensions jointly in a single generation pass, and stands alone as a complete NR assessor whenever no reference is available. 
Stage~2 brings the reference in: it treats a captured training view used for reconstruction as a weak reference. Supervised by human scores, it learns a cue that encourages higher-rated renderings of each scene to have greater embedding similarity to the reference, without requiring pixel-level alignment. 
Stage~3 makes the reference optional rather than required: a bounded gate folds the cue into the frozen regressor to refine its calibrated scores when a reference is available and reverts to the Stage-1 prediction otherwise, so a single model supports both the RR and NR assessment.

On 3DGS-VBench+, 3DGSV-Assessor surpasses traditional metrics, deep VQA models, and recent LMM-based assessors on all three dimensions, and it transfers zero-shot to other NVS quality assessment benchmarks.

The main contributions of this work are summarized as follows:
\begin{itemize}
    \item We construct \textbf{3DGS-VBench+}, a large-scale, multi-dimensional subjective VQA benchmark for compressed 3DGS, comprising 3{,}036 rendered videos across 33 scenes and 6 compression algorithms, annotated with MOSs for overall, geometry, and color dimensions. Extending the preliminary conference version 3DGS-VBench~\cite{3DGS-VBench}, this work adds 22 synthetic scenes to the original 11 real-world scenes, refines the distortion-level design for each algorithm, and expands the single overall dimension into three fine-grained dimensions. 
    These extensions grow the benchmark from 660 to 3,036 models and their rendered videos. The annotation volume grows correspondingly, from 660 to 9,108 MOSs.
    \item We propose \textbf{3DGSV-Assessor}, the first multi-dimensional, reference-flexible VQA metric tailored to 3DGS. This LMM-based framework jointly scores the overall, geometry, and color dimensions, and exploits a captured view of the scene as a weak reference through a bounded gating mechanism, supporting both RR and NR assessment in a single model.
    \item Extensive experiments show that 3DGSV-Assessor achieves SOTA correlation with human perception on 3DGS-VBench+ under both RR and NR settings, and generalizes well to other NVS quality assessment benchmarks.

\end{itemize}

%% file: sec/2_related_work.tex
\section {Related Work}

\begin{table*}[!]
\centering
\vspace{-5mm}
\caption{Summary of existing NVS quality evaluation datasets, with ``Syn'' and ``Real'' representing synthetic and real scenes. }
\vspace{-1mm}
\renewcommand\arraystretch{0.85}
\label{tab:relate}
\resizebox{1\textwidth}{!}{\begin{tabular}{ccccccccccccc}
\hline
Dataset & \multirow{2}{*}{Name} & \multirow{2}{*}{Year}  
&\multicolumn{2}{c}{Scenes} & 
\multicolumn{2}{c}{NVS Models} 
& Render
& Distortion 
&\multirow{2}{*}{Annotation}
& Number of & Number of &  \multirow{2}{*}{MOSs/JODs}
\\

 Type & & & Syn & Real & NeRF & 3DGS & Mode & Level design & & NeRF/3DGS &videos/imgs \\
 \hline

\multirow{2}{*}{IQA} & NVS-QA\cite{NVS-QA} & 2025 & - & 13 & 2 & 3 & 1 viewpoint& \xmark & MOS & 65 & 65 & 65 \\
& 3DGS-IEval-15K+ \cite{3DGSI-Assessor} & 2026 & - & 10 & - & 6 & 20 viewpoints
 & \cmark & MOS & 760 & 15,200 & 45,600\\

\cdashline{1-13}[7pt/5pt]  
\multirow{7}{*}{VQA} & NeRF-QA\cite{NeRF-QA}  & 2023 & 4 & 4 & 7 & - &$360^{\circ}$ & \xmark & DMOS & 48 & 48 &48\\
& NeRF-VSQA\cite{NeRF-VSQA} & 2024 & 8 & 8  & 7 & - & $360^{\circ}$+Front & \xmark & DMOS & 88 & 88 & 88 \\
& FFV\cite{FFV} & 2024 & - & 22 & 8 & - & \textit{Front} & \xmark & Pairs & 220 & 220 & 144 \\
& ENeRF-QA\cite{ENeRF-QA} & 2024 & 22 & - &  4 & - & $360^{\circ}$ & \cmark & MOS & 440 & 440 & 440 \\
& GSC-QA\cite{GSC-QA} & 2024 & 9 & 6 & - & 1 & $360^{\circ}$ & \cmark & MOS & 120 & 120 & 120 \\
& GS-QA \cite{GS-QA}& 2025 & - & 8 & - & 7 & $360^{\circ}$ + Front & \xmark & DMOS & 64 & 64 & 64 \\
& NVS-QA\cite{NVS-QA} & 2025 & - & 13 & 2 & 3 & $360^{\circ}$+Front & \xmark & MOS & 65 & 65 & 65 \\
& 3DGS-VBench(Ours)\cite{3DGS-VBench} & 2025 & - & 11 & - & 6 & $360^{\circ}$ & \cmark & MOS & 660 & 660 & 660 \\
& \textbf{3DGS-VBench+(Ours)} & \textbf{2026} & \textbf{22} & \textbf{11} & - & \textbf{6} & \textbf{$360^{\circ}$} & \cmark & \textbf{MOS} & \textbf{3,036} & \textbf{3,036} & \textbf{9,108} \\
\hline
\end{tabular}}
\vspace{-2mm}
\end{table*}

\subsection{Quality Assessment Databases for NVS}
Quality assessment research for NVS spans both image-level and video-level evaluation, as summarized in Table~\ref{tab:relate}. On the IQA side, NVS-QA \cite{NVS-QA} introduces a benchmark covering NeRF and 3DGS pipelines with 65 image samples. 3DGS-IEval-15K+ \cite{3DGSI-Assessor,3DGS-IEval-15K} establishes the first large-scale 3DGS IQA dataset comprising 15,200 images rendered from 10 real-world scenes across 6 representative compression algorithms with multi-dimensional MOS annotations.
In parallel, the VQA branch of NVS evaluation shifts from NeRF studies toward 3DGS-centered benchmarks. NeRF-QA \cite{NeRF-QA} and NeRF-VSQA \cite{NeRF-VSQA} provide subjective quality assessment datasets containing 48 and 88 videos, respectively. FFV \cite{FFV} adopts a pairwise comparison protocol over 220 samples. ENeRF-QA \cite{ENeRF-QA} contributes systematic distortion modeling by leveraging NeRF compression pipelines across 440 sequences and characterizing nine NeRF-specific distortion categories. For 3DGS content, GSC-QA \cite{GSC-QA} examines a single 3DGS compression scheme with 120 samples, GS-QA \cite{GS-QA} offers comparative evaluation between 3DGS and NeRF over 64 samples, and NVS-QA \cite{NVS-QA} contributes an additional 65 video samples alongside its image counterpart.
However, existing 3DGS VQA datasets exhibit three key limitations: few systematically design compression parameters to cover diverse distortion types and severity levels; most contain fewer than 500 samples; and all provide only overall MOS, lacking the multi-dimensional annotations necessary for fine-grained video screening and downstream 3DGS optimization.
To address these gaps, we present 3DGS-VBench+, which contains 3{,}036 compressed 3DGS video sequences with three-dimensional MOS annotations covering overall, geometry, and color quality, constituting the largest 3DGS VQA dataset to date and the first to provide multi-dimensional subjective labels for 3DGS video content.

\subsection{Quality Assessment Metrics for NVS}
Many video quality assessment models have been proposed in the literature.
Learning-based VQA models (\textit{e.g.}, VSFA~\cite{VSFA}, SimpleVQA~\cite{simpleVQA}, FAST-VQA~\cite{FAST-VQA}) learn quality-aware representations to predict overall video quality but do not explicitly provide separate geometry and color scores in their original formulations, which is crucial for 3DGS quality monitoring. 
LMM-based models~\cite{Qwen2.5-VL,GLM-4.1V,InternVL3} offer strong visual understanding and the flexibility to report multiple dimensions, yet they remain limited in their sensitivity to low-level perceptual quality and often fail to give precise quality scores.
For NVS in particular, evaluation still leans heavily on traditional IQA metrics (\textit{e.g.}, PSNR, SSIM~\cite{PSNR}), which have been found inadequate in assessing perceptual quality of NVS content~\cite{NVS-SQA,3DGS-VBench,ENeRF-QA,3DGS-IEval-15K}.

Beyond these generic metrics, several NVS-specific quality assessment models have been developed for different evaluation settings.
NVS-SQA~\cite{NVS-SQA} learns quality representations via self-supervised contrastive learning, but targets within-scene relative ranking rather than absolute quality prediction across scenes, and thus does not directly apply where absolute quality must be compared across scenes and methods, for example in application-side quality monitoring. Besides, its contrastive pairs rely on traditional distortions (\textit{e.g.}, noise, blur~\cite{tradition-distortion-1,BRISQUE}) rather than artifacts specific to 3DGS.
NeRF-NQA~\cite{NeRF-NQA} performs no-reference scene-level scoring, but requires a densely sampled view sequence with camera poses and a COLMAP point cloud, which are unavailable in many practical scenarios and in many subjective 3DGS datasets. Its original formulation is further restricted to front-facing NeRF scenes, leaving 360$^{\circ}$ content and 3DGS to future work. We therefore exclude it from the present comparison.
3DGSI-Assessor~\cite{3DGSI-Assessor} is a multi-dimensional metric, but is designed for 3DGS image rather than video assessment.
Motivated by these observations, we propose an LMM-based framework for all-in-one, reference-flexible 3DGS VQA that supports both NR and RR settings.

%% file: sec/3_database.tex
\section{DATABASE CONSTRUCTION}
\begin{figure*}[t]
    \centering
    \vspace{-3mm}
    \includegraphics[width=1\linewidth]{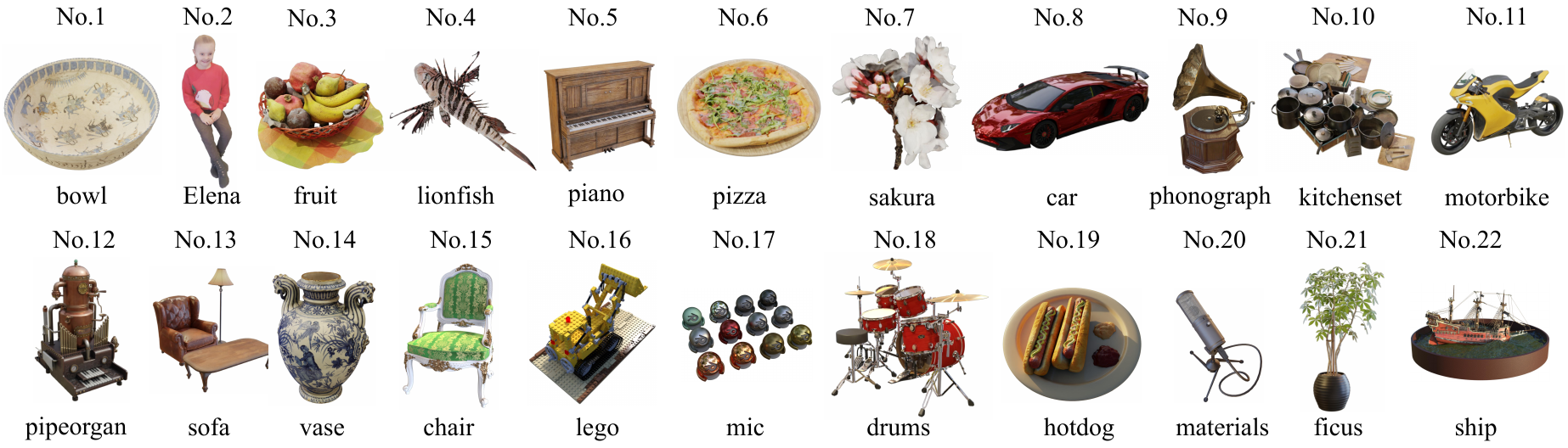}
    \vspace{-7mm}
    \caption{22 selected synthetic source content in 3DGS-VBench+ with diverse material properties.} 
    \label{figure:syn_scene}
    \vspace{-2mm}
\end{figure*}

\begin{figure*}[!t]
    \centering
    \vspace{-1mm}
    \includegraphics[width=1\linewidth]{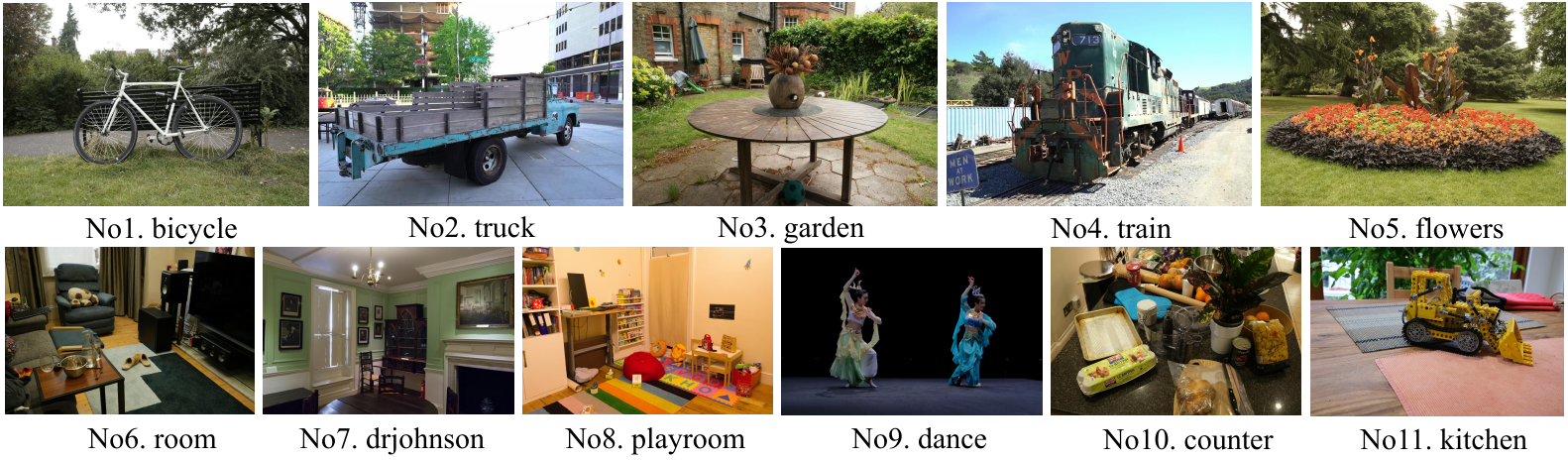}
    \vspace{-7mm}
    \caption{11 selected real source content in 3DGS-VBench+: scenes 1-5 depict outdoor scenes, while scenes 6-11 depict indoor scenes.} 
    \label{figure:real_scene}
    \vspace{-3mm}
\end{figure*}

\subsection{Source Content Selection}
\label{Source_Content_Selection}

To enable comprehensive quality assessment under diverse visual conditions, we collect 33 source scenes comprising 22 synthetic and 11 real-world scenes drawn from multiple canonical NVS multiview datasets.
For synthetic content, as shown in Fig.~\ref{figure:syn_scene} we adopt 22 scenes from two well-established sources: 8 scenes from NeRF\_Synthetic \cite{nerf} (\textit{chair}, \textit{lego}, \textit{mic}, \textit{drums}, \textit{hotdog}, \textit{materials}, \textit{ficus}, \textit{ship}) and 14 scenes from Explicit-NeRF-QA-Synthetic \cite{ENeRF-QA} (\textit{bowl}, \textit{Elena}, \textit{fruit}, \textit{lionfish}, \textit{piano}, \textit{pizza}, \textit{sakura}, \textit{car}, \textit{phonograph}, \textit{kitchenset}, \textit{motorbike}, \textit{pipeorgan}, \textit{sofa}, \textit{vase}).
For real-world content, as shown in Fig.~\ref{figure:real_scene}, we select 11 scenes from four multiview datasets. Two outdoor scenes are obtained from Tanks \& Temples \cite{Tanks-Temples}: \textit{train} (980~$\times$~545) and \textit{truck} (979~$\times$~546). Two indoor scenes, \textit{playroom} (1264~$\times$~832) and \textit{drjohnson} (1332~$\times$~876), are sourced from Deep Blending \cite{Deep-Blending}. Six scenes are obtained from Mip-NeRF360 \cite{Mip-NeRF-360}, including three outdoor scenes: \textit{bicycle} (1237~$\times$~822), \textit{flowers} (1256~$\times$~828), and \textit{garden} (1297~$\times$~840), and three indoor scenes \textit{counter} (1558~$\times$~1038), \textit{kitchen} (1558~$\times$~1039), and \textit{room} (1557~$\times$~1038). We also include a human scene, \textit{Dance\_Dunhuang\_Pair} (1600~$\times$~876), from PKU-DyMVHumans \cite{pku}.

This selection ensures diversity for 3DGS
training and quality assessment. The joint inclusion of synthetic and real content covers two complementary regimes for 3DGS applications. The synthetic subset spans a wide range of object categories, surface materials, geometric structures, and texture complexities. The real-world subset encompasses varied lighting conditions, reflective materials, indoor and outdoor environments, complex textures, and human subjects.
For resolution settings, we maintain native resolution settings from the original 3DGS work \cite{3DGS} for real-world scenes, and adopt the conventional 800~$\times$~800 resolution for synthetic scenes, promoting consistent benchmarking and fair comparison in the NVS community.

\subsection{Camera Trajectory Construction for Video Rendering}
\label{Camera Trajectory Construction for Video Rendering}

Synthetic multi-view image datasets \cite{nerf, ENeRF-QA} are constructed under virtual camera setups, with viewpoints organized into training, testing, and validation sets. The training set is used for 3DGS model training, the testing set provides reference images for quality assessment of reconstructed scenes, and the validation set defines a continuous camera trajectory for rendering video sequences for human viewing. For each synthetic scene, this native validation trajectory comprises 300 viewpoints \cite{ENeRF-QA}, which are used directly for subsequent 3DGS rendering.
However, real-world multi-view image datasets \cite{Mip-NeRF-360,Tanks-Temples,Deep-Blending} are captured with sparsely distributed physical cameras and provide only training and testing sets, lacking the dense, continuous camera paths for smooth video rendering. Consequently, an additional validation viewpoint set must be constructed for each real-world scene to enable consistent video-based subjective evaluation.

To create the validation sets, we first train each 3DGS model on the training set to obtain its point cloud representation, which serves as a spatial reference for camera path design in Blender. For each real-world scene, we construct a smooth $360^{\circ}$ orbital trajectory with the camera parameters adjusted to ensure appropriate framing of the main scene content, and uniformly sample 600 continuous viewpoints along it, consistent with the viewpoint count adopted by GSC-QA \cite{GSC-QA} for real-world content. The resulting validation sets yield smooth video sequences with resolutions consistent with their source datasets, providing a basis for video-level subjective quality assessment.

\vspace{-2pt} 
\subsection{3DGS Model Selection \& Compression Parameter Design}
\label{3DGS Model Selection Compression Parameter Design}

To systematically characterize the distortions introduced by 3DGS generative compression, we select $6$ representative algorithms and, for each, define compression parameter configurations to generate a set of reconstructions covering diverse distortion types and varying intensities.
Based on compression architecture, these 3DGS methods fall into two groups: \textit{single-compression-parameter} methods (HAC \cite{HAC}, Scaffold-GS \cite{Scaffold-GS}), which adopt anchor-based representations where a single dominant compression parameter controls the rate–distortion trade-off; and \textit{multi-compression-parameter} methods (Compact-3DGS \cite{Compact-3DGS}, CompGS \cite{CompGS}, C3dgs \cite{C3dgs}, LightGaussian \cite{LightGaussian}), which independently configure the compression parameters of geometric and color attributes.

For each key compression parameter, we define a set of \textbf{Compression Levels (CL)} that monotonically control distortion intensity. Single-parameter methods assign eight levels ($\mathrm{CL}1$--$\mathrm{CL}8$) to their key parameter, whereas multi-parameter methods assign five levels ($\mathrm{CL}1$--$\mathrm{CL}5$) to each key parameter. A smaller index denotes stronger compression, and the highest level corresponds to the highest-quality configuration.
For real-world scenes, which are the main application scenario of 3DGS, the official methods provide recommended default settings, which we adopt as the highest level. For synthetic scenes, we adopt the official settings where available and otherwise use a high-quality configuration as the highest level.

For each method, we strategically combine the compression parameters to obtain 3DGS samples of varying \textbf{Distortion Levels (DL)} and types. Single-parameter methods simply traverse $(CL1–CL8)$, yielding $\mathbf{8}$ DL per scene. For multi-parameter methods, we combine the per-parameter CLs through three complementary strategies: (1) \textit{Single-parameter ablation}, where one key parameter varies across $(CL1–CL4)$ while the others are fixed at their default optimal level $CL5$.
Since each key parameter mainly governs either geometric or color attributes, this strategy provides a controlled basis for dimension-specific annotation.
(2) \textit{Cross combination}: eight cross configurations per method are constructed by jointly assigning non-default values to two or more key parameters, so that cross samples fill the distortion space between ablation samples rather than repeat them. These configurations produce compound distortions beyond single-parameter perturbation, broadening the coverage of distortion types and better emulating practical deployment configurations, thereby enriching dataset diversity and the generalization of trained VQA models. The eight cross configurations of each method are listed in the supplementary material.
(3) \textit{Default optimal configuration}: an all-CL5 reconstruction that provides the highest-quality reference of each method. Consequently, a method with $n$ key parameters yields \boldmath $4n(ablation) + 8 (cross)+ 1 (default)$ \textbf{DL} per scene.

All methods follow the above combination strategy on both real and synthetic scenes. However, two distinct sets of CL values are designed separately for each content type, as synthetic and real scenes differ in visual characteristics and distortion tolerance, necessitating independent parameter calibration to maintain a rich and comparable perceptual-quality span within each scene type.
The selected 3DGS algorithms and detailed CL configurations are presented as follows:

\paragraph{\textbf{CompGS}~\cite{CompGS}} quantizes Gaussian attributes with $K$-means, maintaining separate codebooks for geometry and color; the two codebook capacities (g-size and c-size), together with an opacity regularization parameter (reg), determine the achievable fidelity. With 3 key parameters, 
\textbf{we generate \boldmath 3 parameters $\times$ 4 distorted CL $=$ 12 single-parameter ablation configurations, 8 cross configurations, and 1 default sample with all optimal parameters, resulting in 21 DL per scene}. CL settings are given in Table \ref{tab:CompGS}.

\begin{table}[H]
\vspace{-2.5mm}
  \caption{CL for compression parameters in \textbf{CompGS}.}
  \label{tab:CompGS}
  \centering
  \vspace{-1.5mm}
  \resizebox{0.49\textwidth}{!}{
  \begin{tabular}{c!{\vrule}c!{\vrule}ccccc}
    \toprule
    \textbf{Type} & \textbf{Parameter} & \textbf{CL1} & \textbf{CL2} & \textbf{CL3} & \textbf{CL4} & \textbf{CL5}
    \\
    \midrule
    \multirow{3}{*}{Real} & \textbf{g-size} & 1 & 2 & $2^3$ & $2^5$ & $2^{12}$
    \\
    & \textbf{c-size} & 1 & 2 & $2^2$ & $2^3$ & $2^{12}$
    \\
    & \textbf{reg} & 4$\times$1e-6 & 3$\times$1e-6 & 2$\times$1e-6 & 1$\times$1e-6 & 1$\times$1e-7
    \\
    \midrule
    \multirow{3}{*}{Syn} & \textbf{g-size} & 1 & $2^2$ & $2^4$ & $2^6$ & $2^{12}$
    \\
    & \textbf{c-size} & 1 & 2 & $2^2$ & $2^5$ & $2^{12}$
    \\
    & \textbf{reg} & 5$\times$1e-6 & 3$\times$1e-6 & 2$\times$1e-6 & 1$\times$1e-6 & 1$\times$1e-7
    \\
    \bottomrule
  \end{tabular}}
\vspace{-1mm}
\end{table}

\paragraph{\textbf{Compact-3DGS}~\cite{Compact-3DGS}} 
compresses the model along three fronts: a learnable mask weighted by $\lambda$ removes redundant Gaussians, geometry attributes are encoded through residual vector quantization parameterized by codebook size and RVQ depth, and view-dependent color is represented by a hash-grid field (hashmap) in place of SH coefficients. With 3 key parameters, we generate {\boldmath $3\times4(ablation) + 8(cross) + 1(default) = \mathbf{21}$} DL per scene. CL settings are given in Table \ref{tab:Compact-3DGS}.

\begin{table}[H]
\vspace{-3mm}
  \caption{CL for compression parameters in \textbf{Compact-3DGS}.}
  \label{tab:Compact-3DGS}
  \centering
  \vspace{-1.5mm}
  \resizebox{0.49\textwidth}{!}{
  \renewcommand{\arraystretch}{1}
  \begin{tabular}{c!{\vrule}c!{\vrule}ccccc}
    \toprule
    \textbf{Type} & \textbf{Parameter} & \textbf{CL1} & \textbf{CL2} & \textbf{CL3} & \textbf{CL4} & \textbf{CL5}
    \\
    \midrule
    \multirow{3}{*}{Real} & \textbf{hashmap} & $2^1$ & $2^2$ & $2^7$ & $2^{12}$ & $2^{19}$
    \\
    & \textbf{(codebook, rvq\_depth)} & ($2^3$, 1) & ($2^4$, 1) & ($2^6$, 1) & ($2^3$, 6) & ($2^6$, 6)
    \\
    & \textbf{lambda} & 0.014 & 0.012 & 0.01 & 0.006 & 0.0005
    \\
    \midrule
    \multirow{3}{*}{Syn} & \textbf{hashmap} & $2^1$ & $2^2$ & $2^4$ & $2^7$ & $2^{16}$
    \\
    & \textbf{(codebook, rvq\_depth)} & ($2^1$, 1) & ($2^2$, 1) & ($2^4$, 1) & ($2^2$, 6) & ($2^6$, 6)
    \\
    & \textbf{lambda} & 0.022 & 0.020 & 0.018 & 0.015 & 0.004
    \\
    \bottomrule
  \end{tabular}}
\vspace{-1mm}
\end{table}

\paragraph{\textbf{LightGaussian}~\cite{LightGaussian}} 
first removes Gaussians of low global significance, then distills high-degree SH coefficients into lower degrees and vector-quantizes the result; the pruning ratio (prune) and the SH quantization setting (c-ratio, c-size) jointly govern the compression. With 2 key parameters, we generate {\boldmath $2\times4(ablation) + 8(cross) + 1(default) = \mathbf{17}$} DL per scene. CL settings are given in Table \ref{tab:LightGaussian}.

\begin{table}[H]
\vspace{-1mm}
  \caption{CL for parameters in \textbf{LightGaussian}.}
  \label{tab:LightGaussian}
  \centering
  \vspace{-1.5mm}
  \resizebox{0.48\textwidth}{!}{
  \begin{tabular}{c!{\vrule}c!{\vrule}ccccc}
    \toprule
    \textbf{Type} & \textbf{Parameter} & \textbf{CL1} & \textbf{CL2} & \textbf{CL3} & \textbf{CL4} & \textbf{CL5}
    \\
    \midrule
    \multirow{2}{*}{Real} & \textbf{prune} & 0.97 & 0.95 & 0.90 & 0.85 & 0.66
    \\
    & \textbf{(c-ratio, c-size)} & (1, 2) & (1, $2^2$) & (1, $2^3$) & (1, $2^5$) & (0.6, $2^{13}$)
    \\
    \midrule
    \multirow{2}{*}{Syn} & \textbf{prune} & 0.95 & 0.90 & 0.85 & 0.75 & 0.66
    \\
    & \textbf{(c-ratio, c-size)} & (1, 1) & (1, 2) & (1, $2^3$) & (1, $2^5$) & (0.6, $2^{12}$)
    \\
    \bottomrule
  \end{tabular}}
\vspace{-1mm}
\end{table}

 \paragraph{\textbf{C3dgs}~\cite{C3dgs}} 
clusters geometry and color attributes into codebooks and entropy-codes them; for each attribute, a codebook size paired with an importance threshold: (c-size, c-include) for color and (g-size, g-include) for geometry, decides which primitives are retained versus quantized. With 2 key parameters, we generate {\boldmath $2\times4(ablation) + 8(cross) + 1(default) = \mathbf{17}$} DL per scene. CL settings are given in Table \ref{tab:c3dgs}.

\begin{table}[H]
\vspace{-2.5mm}
  \caption{CL for compression parameters in \textbf{c3dgs}.}
  \label{tab:c3dgs}
  \centering
  \vspace{-1.5mm}
  \resizebox{0.49\textwidth}{!}{
  \renewcommand{\arraystretch}{1}
  \begin{tabular}{c!{\vrule}c!{\vrule}ccccc}
    \toprule
    \textbf{Type} & \textbf{Parameter} & \textbf{CL1} & \textbf{CL2} & \textbf{CL3} & \textbf{CL4} & \textbf{CL5 (default)}
    \\
    \midrule
    \multirow{2}{*}{Real} & \textbf{(c-size, c-include)} & (1, 0.6) & (2, 0.6) & ($2^2$, 0.6) & ($2^5$, 0.6) & ($2^{12}$, 0.6$\times$1e-6) 
    \\
    & \textbf{(g-size, g-include)} & (1, 0.3) & ($2^3$, 0.3) & ($2^5$, 0.3) & ($2^8$, 0.3) & ($2^{12}$, 0.3$\times$1e-5)
    \\
    \midrule
    \multirow{2}{*}{Syn} & \textbf{(c-size, c-include)} & (1, 0.6) & (2, 0.6) & ($2^2$, 0.6) & ($2^4$, 0.6) & ($2^{12}$, 0.6$\times$1e-6) 
    \\
    & \textbf{(g-size, g-include)} & (1, 0.3) & ($2^2$, 0.3) & ($2^4$, 0.3) & ($2^6$, 0.3) & ($2^{12}$, 0.3$\times$1e-5)
    \\
    \bottomrule   
    
  \end{tabular}}
\vspace{-1mm}
\end{table}

\paragraph{\textbf{Scaffold-GS}~\cite{Scaffold-GS}} 
organizes Gaussians around anchor points placed on a voxel grid and derives local Gaussians from these anchors; the voxel size (vsize) sets the anchor density and thereby the compression strength. As shown in Table \ref{tab:Scaffold-GS}, with $1$ compression parameter corresponding to $8$ CL, we generate \boldmath$8$ DL per scene.

\begin{table}[H]
\vspace{-2.5mm}
  \caption{CL for compression parameter in \textbf{Scaffold-GS}.}
  \label{tab:Scaffold-GS}
  \centering
  \vspace{-1.5mm}
  \resizebox{0.48\textwidth}{!}{
  \begin{tabular}{c!{\vrule}cccccccc}
    \toprule
    \textbf{Param} & \textbf{CL1} & \textbf{CL2} & \textbf{CL3} & \textbf{CL4} & \textbf{CL5} & \textbf{CL6} & \textbf{CL7} & \textbf{CL8}
    \\
    \midrule
    \textbf{vsize (real)} & 0.350 & 0.250 & 0.200 & 0.160 & 0.120 & 0.080 & 0.050 & 0.001
    \\
    \midrule
    \textbf{vsize (syn)} & 0.350 & 0.250 & 0.180 & 0.120 & 0.080 & 0.040 & 0.015 & 0.001
    \\
    \bottomrule
  \end{tabular}}
\vspace{-1mm}
\end{table}

\paragraph{\textbf{HAC}~\cite{HAC}} 
couples an anchor-based representation with a hash-grid-assisted context model for entropy coding, and a single trade-off weight $\lambda$ regulates the balance between bitrate and reconstruction fidelity. As shown in Table \ref{tab:HAC}, with one compression parameter has $8$ CLs, we generate \boldmath$8$ DL per scene.

\begin{table}[H]
\vspace{-2mm}
  \caption{CL for compression parameter in \textbf{HAC}.}
  \label{tab:HAC}
  \centering
  \vspace{-1.5mm}
  \resizebox{0.48\textwidth}{!}{
  \begin{tabular}{c!{\vrule}cccccccc}
    \toprule
    \textbf{Param} & \textbf{CL1} & \textbf{CL2} & \textbf{CL3} & \textbf{CL4} & \textbf{CL5} & \textbf{CL6} & \textbf{CL7} & \textbf{CL8}
    \\
    \midrule
    \textbf{$\lambda$ (real)} & 0.800 & 0.600 & 0.400 & 0.300 & 0.200 & 0.120 & 0.060 & 0.004
    \\
    \midrule
    \textbf{$\lambda$ (syn)} & 8.000 & 4.000 & 0.400 & 0.250 & 0.150 & 0.070 & 0.030 & 0.004
    \\
    \bottomrule
  \end{tabular}}
\vspace{-1mm}
\end{table}

In all, we trained ($\mathbf{11}$ real $+$ $\mathbf{22}$ synthetic) scenes $\times$ $\mathbf{(21 + 21 + 17 + 17 + 8 + 8)}$ model settings $\mathbf{= 33}$ scenes $\mathbf{\times 92}$ reconstructions $\mathbf{= 3036}$ 3DGS reconstruction models, covering a broad and well-controlled space of distortion types and intensities.

\subsection{Subjective Experiment and Data Processing}
\label{Subjective_Experiment_and_Data_Processing}
\subsubsection*{\bf Subjective Experiment Setup} 

\begin{figure*}[!t]
    \centering
    \vspace{-5mm}
    \includegraphics[width=1\linewidth]{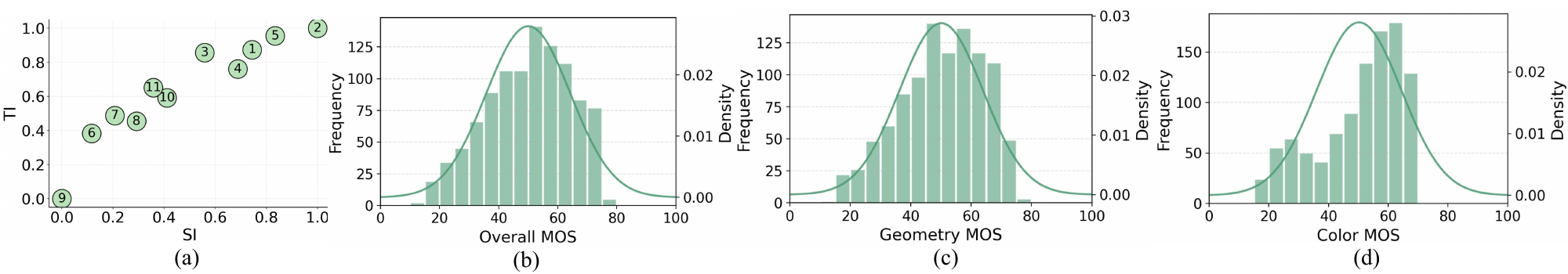}
    \vspace{-7.5mm}
    \caption{Statistical analysis of the 11 real-world scenes in 3DGS-VBench+ from content complexity and quality perspectives. (a) Scene distribution across SI and TI dimensions. (b)-(d) MOS distribution for overall quality, geometry quality, and color quality.}
    \label{fig:real_dataset_exp}
    \vspace{-1mm}
\end{figure*}

\begin{figure*}[!t]
    \centering
    \vspace{-1mm}
    \includegraphics[width=1\linewidth]{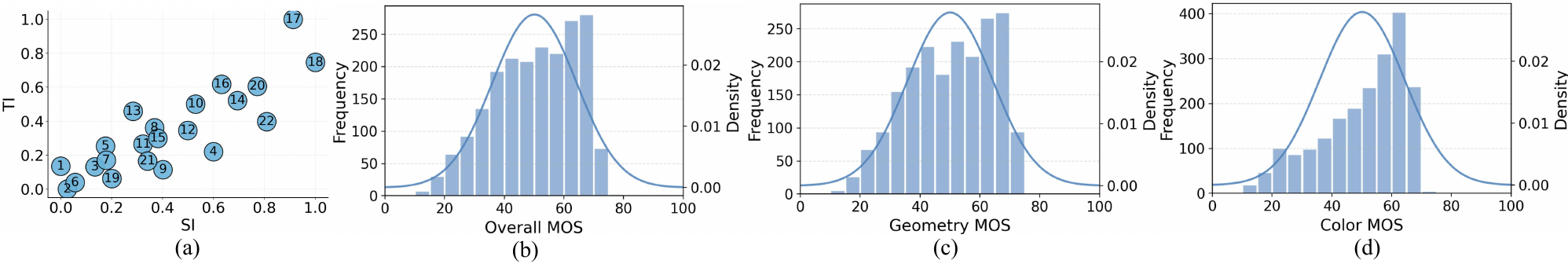}
    \vspace{-7.5mm}
    \caption{Statistical analysis of the 22 synthetic scenes in 3DGS-VBench+ from content complexity and quality perspectives. (a) Scene distribution across SI and TI dimensions. (b)-(d) MOS distribution for overall quality, geometry quality, and color quality.} 
    \label{fig:syn_dataset_exp}
    \vspace{-1.5mm}
\end{figure*}

Each of the 3{,}036 reconstructed 3DGS models is first turned into a Processed Video Sequence (PVS) by rendering along its scene's validation trajectory (Sec.~\ref{Camera Trajectory Construction for Video Rendering}). The frames are encoded using FFMPEG with the libx265 codec at 30 fps and a constant rate factor of 10, yielding 3{,}036 PVSs in total.
To characterize the perceptual effects of 3DGS compression beyond a single overall score, each PVS is rated along three dimensions: (1) \textit{Overall Quality} for global visual fidelity, (2) \textit{Geometry Quality} for structural integrity and spatial accuracy, and (3) \textit{Color Quality} for chromatic accuracy and plausibility. Every PVS is presented individually and scored on the 11-point numerical quality scale recommended by ITU-T P.910 \cite{ITU1}. The playback and rating interface is built with the Python Tkinter toolkit, and all sessions run on a calibrated 27-inch AOC Q2790PQ monitor in a controlled laboratory under fixed standard illumination. Finally, we obtain a total of 136{,}620 reliable score ratings (15 ratings $\times$ 3 dimensions $\times$ 3{,}036 PVSs).

\subsubsection*{\bf Subjective Data Processing} 

In order to obtain the MOS scores for each video, we first convert the ratings into Z-scores, and then linearly remapped to the [0, 100] range, as demonstrated below:

\begin{equation}
z_{ij}^{} =\frac{r_{ij} - \mu_{i}  }{\sigma _{i} } ,   z_{ij}^{'} = \frac{100(z_{ij}+3) }{6} 
\end{equation}

\begin{equation}
\mu _{i}=\frac{1}{N_{i} }\sum_{j=1}^{N_{i} }  r_{ij},  \sigma _{i}=\sqrt{\frac{1}{N_{i}-1 }\sum_{j=1}^{N_{i} }(r_{ij} - \mu _{i})^{2}     },
\end{equation}
where $r_{ij}$ is the raw score given by the $i$-th subject to the $j$-th PVS. $N_{i}$ is the number of PVSs rated by subject $i$. The MOS of each PVS is then obtained as the mean of the rescaled z-scores across all subjects as follows:

\vspace{-1mm}
\begin{equation}
MOS_{j}=\frac{1}{M_j}\sum_{i=1}^{M_j}z_{ij}^{'},
\end{equation}
where $M_j$ denotes the number of valid subjects who rated the $j$-th PVS, and $z_{ij}^{'}$ refers to the rescaled z-score. Therefore, applied independently to the three dimensions, a total of 9,108 MOSs (3 dimensions $\times$ 3,036 PVSs) are obtained.

\subsection{Subjective Data Analysis}

\subsubsection*{\bf Diversity of Source Content}

\begin{figure*}[!t]
    \centering
    \vspace{-3mm}
    \includegraphics[width=1\linewidth]{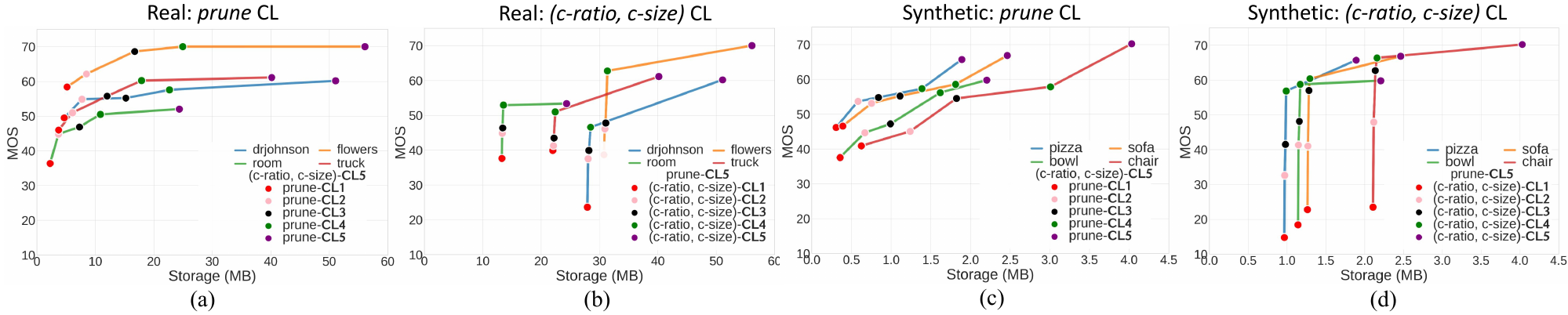}
    \vspace{-7.5mm}
    \caption{MOS vs. Storage (MB) for LightGaussian on representative real-world and synthetic scenes in 3DGS-VBench+. (a) 5 CLs of \emph{prune} for real-world scenes. (b) 5 CLs of \emph{(c-ratio, c-size)} for real-world scenes. (c) 5 CLs of \emph{prune} for synthetic scenes. (d) 5 CLs of \emph{(c-ratio, c-size)} for synthetic scenes.}
    \label{fig:mos-size}
    \vspace{-2mm}
\end{figure*}

To verify the diversity of the source content, we analyze the Spatial
Information (SI) \cite{ITU1} and Temporal Information (TI)
\cite{ITU1} characteristics of the selected scenes. Fig.~\ref{fig:real_dataset_exp}(a)
reports the SI-TI distribution of the 11 real-world scenes, where each
numbered point corresponds to the scene index in Fig.~\ref{figure:real_scene}.
Fig.~\ref{fig:syn_dataset_exp}(a) reports the SI-TI distribution for the 22 synthetic
scenes, with the numbered points corresponding to the scene index in
Fig.~\ref{figure:syn_scene}. Both scene types yield balanced distribution across both complexity dimensions, confirming the visual diversity of our selected scenes and enhancing the generalization of our dataset for quality assessment model training.

\subsubsection*{\bf MOS Analysis}
To assess the rationality of the collected scores, Fig.~\ref{fig:real_dataset_exp}(b)-(d) and Fig.~\ref{fig:syn_dataset_exp}(b)-(d) plot the MOS distributions of overall, geometry, and color quality for the real-world and synthetic scenes, respectively. All three perceptual dimensions show broad quality coverage under both scene types, ensuring sufficient characterization of different distortion levels and balanced data for training quality assessment models. The color quality MOS is  right-skewed toward the high-score region ($>$60), since the anchor-based methods (HAC and Scaffold-GS) predominantly affect geometric quality under compression.

\subsubsection*{\bf CL Design Analysis}
To verify the collected MOS and the designed compression levels, we take LightGaussian as a representative case and inspect the MOS--storage relationship over the CL settings listed in Table~\ref{tab:LightGaussian}. As noted in Section~\ref{3DGS Model Selection Compression Parameter Design}, the CL values are calibrated separately for the two content types, so we select four scenes from each as examples to validate both CL grids: \textit{truck} (Tanks \& Temples), \textit{drjohnson} (Deep Blending), \textit{room} and \textit{flowers} (Mip-NeRF360) for the real-world type, and \textit{pizza}, \textit{sofa}, \textit{bowl} and \textit{chair} for the synthetic type. Fig.~\ref{fig:mos-size}(a) and (c) traverse the CLs of \textbf{\textit{prune}} on the real and synthetic scenes respectively, while Fig.~\ref{fig:mos-size}(b) and (d) traverse the CLs of \textbf{\textit{(c-ratio, c-size)}}, with the remaining parameter fixed at its default level CL5.

In all four panels, MOS decreases monotonically as the CL index descends from CL5 to CL1 with the model compressed to a smaller size. Since stronger compression discards more information, this consistency supports both the reliability of the collected MOS and the validity of the CL grids designed for each content type.

Both parameters show that in most cases the default configuration CL5 retains a varying degree of redundancy: CL5 - CL4 substantially reduces the storage at a relatively small MOS cost. The two parameters then diverge. \textbf{\textit{prune}} keeps trading storage for quality gradually, whereas CL4 - CL1 of \textbf{\textit{(c-ratio, c-size)}} remain within a narrow storage interval while quality falls sharply, sacrificing perceptual fidelity almost without additional saving. \textbf{\textit{Prune}} therefore offers a more favorable quality-compression trade-off in practical use.

The content types also differ. On real-world scenes, compressing \textbf{\textit{prune}} to CL3 is nearly free and only lower levels incur clear penalties, while synthetic scenes decline steadily from the first step, as their far fewer Gaussians mean an identical pruning ratio removes more of the primitives carrying visible structure. Compression behavior is therefore not transferable between content types, supporting both their joint coverage and their independent CL calibration.

%% file: sec/4_method-6.tex
\section{The 3DGSV-Assessor Approach}
\label{sec:method}

\begin{figure*}[!t]
    \centering
    \vspace{-3mm}
    \includegraphics[width=1\linewidth]{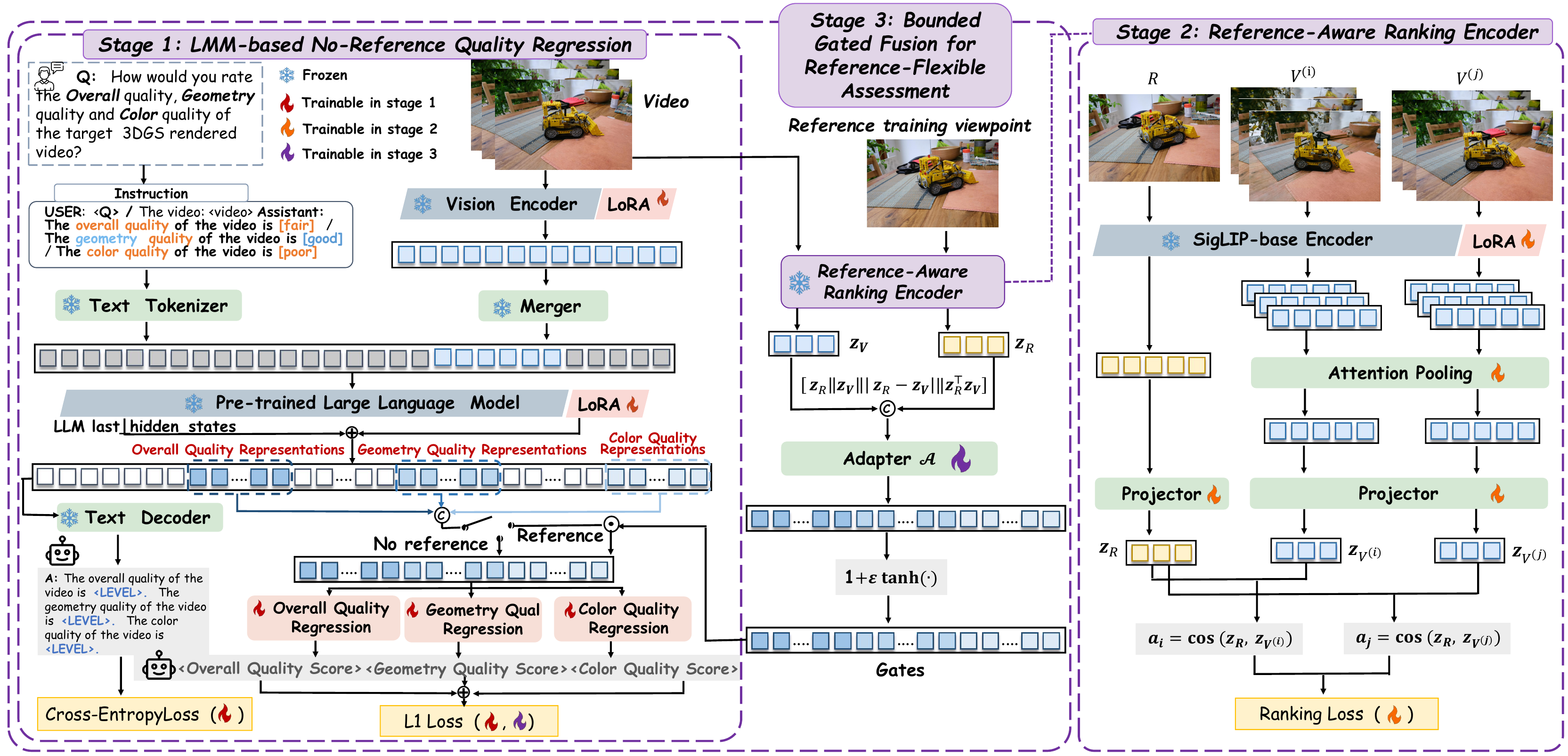}
    \vspace{-6.5mm}
    \caption{Framework of 3DGSV-Assessor, an LMM-based method for all-in-one, reference-flexible 3DGS VQA. Left: Stage~1 is an LMM-based no-reference regressor that scores all three dimensions jointly. Right: Stage~2 encodes a captured view and same-scene videos into a reference-agreement cue, trained by ranking to pull higher-rated renderings closer to the reference. Middle: Stage~3 turns the cue into a bounded gate that refines the Stage-1 representation when a reference is available and is bypassed otherwise, so a single model serves both RR and NR assessment.}
    \label{fig:method}
    \vspace{-2.5mm}
\end{figure*}

In this section we present \textbf{3DGSV-Assessor}, an LMM-based framework for all-in-one, reference-flexible VQA of compressed 3DGS. Given a 3DGS-rendered orbital video $V$ and, when available, a captured training viewpoint reference of the scene, the framework outputs a triplet of continuous quality scores $\mathbf{y}^{\star}=(y_o^{\star},y_g^{\star},y_c^{\star})$, where $\mathcal{D}=\{o,g,c\}$ indexes the overall, geometry, and color dimensions. As shown in Fig.~\ref{fig:method}, the framework progresses from a no-reference base to reference-flexible assessment in three stages. Stage~1 (Sec.~\ref{sec:stage1}) is an LMM-based no-reference regressor that scores all three dimensions jointly and by itself constitutes the standalone NR mode. Stage~2 (Sec.~\ref{sec:stage2}) encodes a captured view that accompanies the 3DGS model into a reference-agreement cue, learning to pull higher-rated renderings of each scene closer to its own reference. Stage~3 (Sec.~\ref{sec:stage3}) folds this cue into the frozen regressor through a bounded gate, refining the calibrated scores when a reference is available and reverting exactly to the Stage-1 prediction otherwise, so a single model serves both the RR and the NR setting.

\subsection{Stage 1: LMM-based No-Reference Quality Regression}
\label{sec:stage1}

\textbf{Structure.} We adopt Qwen3.5 (9B)~\cite{Qwen3.5} as the pre-trained vision-language backbone $\mathcal{F}_\theta$, with final-layer width $D_h$. The video $V$ is sampled into a fixed number of $T$ uniformly spaced frames and tokenized by the backbone's native visual encoder. The visual sequence is interleaved with a fixed instruction $I$ requesting a joint verdict, \textit{``The video: \texttt{<video>}. How would you rate the overall quality, geometry quality, and color quality of the video?''}, and the model is supervised to answer with a template in which each dimension is summarized by a quality-level word, \textit{``The overall quality of the video is $\langle w_o\rangle$. The geometry quality of the video is $\langle w_g\rangle$. The color quality of the video is $\langle w_c\rangle$.''}
The training label for each slot $\langle w_d\rangle$ is obtained by mapping the ground-truth MOS $y_d$ to one of five levels through uniform interval division, following ITU~\cite{ITU3}. 
Let $y_d^{\min}$ and $y_d^{\max}$ denote the minimum and maximum MOS of dimension $d$ in the dataset, and $\Delta_d=y_d^{\max}-y_d^{\min}$. We partition $[y_d^{\min}, y_d^{\max}]$ into five equal intervals:
\begin{equation}
w_d=s_k \ \text{ if } \ y_d^{\min}+\tfrac{k-1}{5}\Delta_d<y_d\le y_d^{\min}+\tfrac{k}{5}\Delta_d,
\end{equation}
where $\{s_k\}_{k=1}^{5}=\{\text{Bad},\text{Poor},\text{Fair},\text{Good},\text{Excellent}\}$, and $y_d=y_d^{\min}$ is assigned to $s_1$.

Since a five-level verdict is too coarse to serve as a continuous MOS, we further read the score from the internal representation that produces the verdict rather than from the decoded text. Letting $\mathbf{H}\in\mathbb{R}^{L\times D_h}$ denote the final-layer hidden states over the $L$-token answer, we extract the states at the three positions $\{\pi_o,\pi_g,\pi_c\}$ that predict the quality-level words,
\begin{equation}
\mathbf{h}_d=\mathbf{H}[\pi_d],\qquad d\in\mathcal{D}.
\end{equation}

The three states are concatenated into a shared representation $\mathbf{u}=[\mathbf{h}_o\,\Vert\,\mathbf{h}_g\,\Vert\,\mathbf{h}_c]\in\mathbb{R}^{3D_h}$ and mapped to scores by three independent five-layer MLP heads,
\begin{equation}
\hat{y}_d=\phi_d(\mathbf{u}),\qquad d\in\mathcal{D}.
\end{equation}
Sharing $\mathbf{u}$ preserves cross-dimensional dependencies inherent to 3DGS, while the separate heads grant each dimension a dimension-specific regression function.

\textbf{Training.} 
The backbone is adapted with LoRA~\cite{LORA} on both its visual encoder and language model, leaving the pre-trained weights frozen, while the heads are trained from scratch. Optimization couples a cross-entropy loss $\mathcal{L}_{\mathrm{ce}}$, the autoregressive objective over the answer tokens $\{t_n\}_{n=1}^{L}$,
\begin{equation}
\mathcal{L}_{\mathrm{ce}}=-\frac{1}{L}\sum_{n=1}^{L}\log P\big(t_n\mid t_{<n},V,I;\theta\big),
\end{equation}
with an $\ell_1$ regression loss over dimensions,
\begin{equation}
\mathcal{L}_{\mathrm{reg}}=\sum_{d\in\mathcal{D}}\big|\hat{y}_d-y_d\big|,
\end{equation}
combined as $\mathcal{L}_{\mathrm{base}}=\alpha\,\mathcal{L}_{\mathrm{ce}}+\beta\,\mathcal{L}_{\mathrm{reg}}$, with $\beta$ ramped from zero so the randomly initialized heads do not destabilize early training, as detailed in Sec.~\ref{sec:exp_setup}.

\subsection{Stage 2: Reference-Aware Ranking Encoder}
\label{sec:stage2}
Stage~2 learns a cue measuring how well the rendered video agrees with a reference: a captured training view of the scene (a photograph for real-world scenes and a ground-truth rendering for synthetic ones). 
This setting is practical because such a captured training view inherently exists for any 3DGS model and can accompany it at little extra cost.

\textbf{Structure.}
We build the ranking encoder $\mathcal{G}$ on a SigLIP visual backbone~\cite{siglip} adapted with LoRA and shared between the reference and video branches. For each scene we build a single reference anchor by encoding one such view $R$ and projecting it with a head $p_R$ to an $m$-dimensional unit-norm embedding,
\begin{equation}
\mathbf{z}_R=\overline{p_R\!\big(\mathcal{G}(R)\big)}\in\mathbb{R}^{m},
\end{equation}
where $\overline{(\cdot)}$ denotes $\ell_2$ normalization. 
We use one reference view as the default scene anchor and evaluate the effect of additional views in Sec.~\ref{app:abl_num_refs}. The anchor provides partial appearance information without requiring pixel-level alignment with the rendered video.
The video is summarized from $T'$ uniformly sampled frames via attention pooling and projected with a head $p_V$,
\begin{equation}
\mathbf{z}_V=\overline{p_V\!\big(\mathrm{AttnPool}(\{\mathcal{G}(V_t)\}_{t=1}^{T'})\big)}\in\mathbb{R}^{m}.
\end{equation}

\textbf{Training.}
The cue is trained only to order renderings within a scene. Since such an ordering depends solely on the relative comparison of same-scene pairs, the objective tolerates both noise in the absolute MOS labels and the viewpoint mismatch between a static reference and a moving camera, rewarding scene-level reference agreement rather than pixel-level alignment.
A scene-stratified sampler draws each mini-batch from a single scene spanning its quality levels, so all videos $\mathbf{V}^{(i)}$ in a batch share one anchor $\mathbf{z}_R$. We form the set $\mathcal{P}=\{(i,j)\,:\,y_o^{(i)}-y_o^{(j)}\ge\tau\}$ of pairs whose overall-MOS gap clears a noise margin $\tau$ (a batch with $\mathcal{P}=\emptyset$ contributes zero loss), and apply a margin-ranking penalty (margin $\rho$) on each sample's cosine agreement with the anchor, $a_i=\cos(\mathbf{z}_{\mathbf{V}^{(i)}},\mathbf{z}_R)$,
\begin{equation}
\mathcal{L}_{\text{rank}}=\frac{1}{\max(|\mathcal{P}|,1)}\sum_{(i,j)\in\mathcal{P}}\max\!\big(0,\; \rho-(a_i-a_j)\big),
\end{equation}
driving higher-quality videos closer to the reference. We rank by overall MOS alone, since it is the holistic verdict a subject reaches after integrating every visible degradation, and leave Stage~3 to resolve the cue into per-dimension corrections.
After convergence, $\mathcal{G}$ is frozen and the embeddings $\{\mathbf{z}_R,\mathbf{z}_V\}$ are pre-computed and cached offline, which fixes the cue as a stable input to Stage~3 and removes the encoding cost from its fine-tuning. Since $\mathcal{G}$ is trained only on the training split of 3DGS-VBench+'s video--MOS samples, the cached embeddings introduce no evaluation leakage.

\subsection{Stage 3: Gated Fusion for Reference-Flexible Assessment}
\label{sec:stage3}

Stage~2 yields the reference cue but leaves it detached from the calibrated Stage-1 regressor. Stage~3 bridges the two while keeping the Stage-1 system intact. The reference never enters the LMM context and no Stage-1 weight is updated; the cue acts only through a bounded gate applied in feature space, so the RR and NR modes share an identical generation pass, and withholding the cue recovers the Stage-1 prediction exactly. This non-invasive fusion is what lets a single model serve both settings.

\textbf{Structure.}
From the cached embeddings, we construct a relational cue that combines reference and video features with their difference and similarity:
\begin{equation}
\mathbf{c}=\big[\,\mathbf{z}_R\,\Vert\,\mathbf{z}_V\,\Vert\,|\mathbf{z}_R-\mathbf{z}_V|\,\Vert\,\mathbf{z}_R^{\!\top}\mathbf{z}_V\,\big],
\end{equation}
concatenating the two embeddings, their element-wise absolute difference, and their inner product (the cosine agreement $a$ of Sec.~\ref{sec:stage2}, since both embeddings are unit-norm). A lightweight adapter $\mathcal{A}$, consisting of a LayerNorm followed by a two-layer MLP with GELU activation, maps this cue to a bounded multiplicative modulation of the Stage-1 quality representation,
\begin{equation}
\boldsymbol{\gamma}=\mathbf{1}+\epsilon\tanh\!\big(\mathcal{A}(\mathbf{c})\big)\in[1-\epsilon,\,1+\epsilon]^{3D_h},\qquad
\tilde{\mathbf{u}}=\mathbf{u}\odot\boldsymbol{\gamma},
\end{equation}
after which the frozen heads yield the scores $\tilde{y}_d=\phi_d(\tilde{\mathbf{u}})$. The modulation is centered at $\boldsymbol{\gamma}=\mathbf{1}$ and symmetric about it, so the cue can amplify or suppress any feature, while $\epsilon$ bounds the per-feature scaling, $\|\boldsymbol{\gamma}-\mathbf{1}\|_\infty\le\epsilon$.

\textbf{Training.} Stage~3 freezes the entire backbone, both LoRA adapters, and the reused Stage-1 heads, and trains only the adapter $\mathcal{A}$ with $\mathcal{L}_{\text{reg}}$ computed on the scores $\tilde{y}_d$.
When a scene provides no reference, the cue cannot be formed; $\mathcal{A}$ is then bypassed and $\boldsymbol{\gamma}$ is set to $\mathbf{1}$, so that $\tilde{\mathbf{u}}=\mathbf{u}$. The frozen Stage-1 heads therefore return exactly the Stage-1 NR prediction $\hat{y}_d$, and reference enhancement is optional rather than required.
The scores $\tilde{y}_d$ (RR), or $\hat{y}_d$ (NR) when no reference is available, constitute the framework output $\mathbf{y}^{\star}$, produced in a single generation pass.

%% file: sec/5_exp_opti.tex
\section{Experiments}

\begin{table*}[t]
\vspace{-5mm}
\caption{Performance benchmark on 3DGS-VBench+. $\spadesuit$ Traditional Handcrafted QA models, $\heartsuit$ LMM-based models, $\blacklozenge$ Deep learning-based VQA models. Best results are marked in {\red{RED}} and second-best in {\blue{BLUE}}. *Refers to finetuned models.}
\renewcommand{\arraystretch}{1}
\centering
\vspace{-2mm}
\resizebox{0.9\textwidth}{!}{
  \begin{tabular}{l||ccc|ccc|ccc}
    \toprule
    \multicolumn{1}{l}{} &
    \multicolumn{3}{c}{\textbf{Overall Quality}} &
    \multicolumn{3}{c}{\textbf{Geometry Quality}} &
    \multicolumn{3}{c}{\textbf{Color Quality}} \\
    \cmidrule(lr){2-4} \cmidrule(lr){5-7} \cmidrule(lr){8-10} 
    \textbf{Methods / Metrics} & \textbf{SRCC} & \textbf{PLCC} & \textbf{KRCC} & \textbf{SRCC} & \textbf{PLCC} & \textbf{KRCC} & \textbf{SRCC} & \textbf{PLCC} & \textbf{KRCC} \\
    \midrule
    $\spadesuit$ PSNR\cite{PSNR} & 0.5173 & 0.5042& 0.3652& 0.4976 & 0.4843 & 0.3316 & 0.4782 & 0.4691 & 0.3213 \\
    $\spadesuit$ SSIM\cite{PSNR} & 0.5391 & 0.5073 & 0.3845 & 0.5265 & 0.5076 & 0.3815 & 0.4845 & 0.4535 & 0.3364 \\
    $\spadesuit$ LPIPS\cite{LPIPS} & 0.5448 & 0.4927 & 0.3770 & 0.5518 & 0.5033 & 0.3645 & 0.4683 & 0.4103 & 0.3024 \\
    $\spadesuit$ DISTS \cite{DISTS} & 0.7267 & 0.7246 & 0.5016 & 0.7109 & 0.6937 & 0.4976 & 0.6618 & 0.6403 & 0.4545 \\
    $\spadesuit$ VIF\cite{VIF} & 0.4137 & 0.4228 & 0.2949 & 0.4003 & 0.4461 & 0.2842 & 0.3505 & 0.3537& 0.2631 \\
    $\spadesuit$ FSIM\cite{FSIM} & 0.5991 & 0.5837 & 0.4275 & 0.5918 & 0.5754 & 0.4103 & 0.5377 & 0.5116 & 0.3765 \\
    $\spadesuit$ IW-SSIM\cite{IW-SSIM} & 0.6252 & 0.5753 & 0.4178 & 0.6365 & 0.5718 & 0.4264 & 0.5578 & 0.4932 & 0.3294 \\
    $\spadesuit$ MS-SSIM\cite{MS-SSIM} & 0.5237 & 0.4914 & 0.4075 & 0.5115 & 0.4767 & 0.4045 & 0.4613 & 0.4178 & 0.3631 \\
    $\spadesuit$ BRISQUE\cite{BRISQUE} & 0.2428 & 0.2018 & 0.1673 & 0.2796 & 0.2316 & 0.1976 & 0.2031 & 0.1516 & 0.1679 \\
    
    \midrule
    $\heartsuit$ LLaVA-one-vision (7B)\cite{Llava-one-vision} & 0.5040 & 0.4727 & 0.4471 & 0.4450 & 0.5067 & 0.3550 & 0.3727 & 0.3213 & 0.2833 \\

    $\heartsuit$ mPLUG-Owl3 (7B)\cite{mPLUG-Owl3}& 0.5001 & 0.4986 & 0.4231 & 0.5783 & 0.5644 & 0.5342 & 0.4037 & 0.3535 & 0.2931 \\
    $\heartsuit$ InternVL3.5 (38B) \cite{InternVL3.5} & 0.6794 & 0.6795 & 0.5951 & 0.7459 & 0.7625 & 0.6654 & 0.6657 & 0.6683 & 0.5459 \\
    $\heartsuit$ InternVL3.5 (8B) \cite{InternVL3.5} & 0.5337 & 0.5525 & 0.4763 & 0.6236 & 0.6102 & 0.5330 & 0.4911 & 0.5092 & 0.3975 \\
    $\heartsuit$ InternVL3 (8B)\cite{InternVL3} & 0.4404 & 0.4010 & 0.3059 & 0.5202 & 0.4950 & 0.3971 & 0.3021 & 0.3114 & 0.2123 \\
    $\heartsuit$ GLM-4.1V (9B)\cite{GLM-4.1V} & 0.5857 & 0.6019 & 0.4571 & 0.6264 & 0.6388 & 0.5182 & 0.5213 & 0.4789 & 0.3810  \\
    
    $\heartsuit$ Qwen3-VL (32B)\cite{Qwen3-VL} & 0.7571 & 0.7484 & 0.6395 & 0.7749 & 0.7750 & 0.6493 & 0.7249 & 0.6874 & 0.5948 \\
    $\heartsuit$ Qwen3-VL (8B)\cite{Qwen3-VL}
    & 0.6669 & 0.6553 & 0.5698 & 0.7165 & 0.7173 & 0.6183 & 0.6279 & 0.5952 & 0.4830 \\
    $\heartsuit$ Qwen3.5 (35B) \cite{Qwen3.5} & 0.7102 & 0.6922 & 0.5966 & 0.7858 & 0.7744 & 0.6742 & 0.6067 & 0.5627 & 0.4948 \\
    $\heartsuit$ Qwen3.5 (9B) \cite{Qwen3.5} & 0.6715 & 0.6316 & 0.5406 & 0.6643 & 0.6785 & 0.5205 & 0.6133 & 0.5486 & 0.4757 \\
    $\heartsuit$ Llama3.2-Vision (11B)* \cite{llama3.2} & 0.8547 & 0.8549 & 0.6616 & 0.8676 & 0.8583 & 0.6818 & 0.8315 & 0.8237 & 0.6228  \\
    $\heartsuit$ Qwen3.5 (9B)* \cite{Qwen3.5} & 0.8946 & 0.8983 & 0.6919 & 0.8845 & 0.8818 & 0.6727 & 0.8624 & 0.8618 & 0.6737\\
    \midrule
    $\blacklozenge$ DOVER \cite{DOVER}
   & 0.9154 & 0.9093 & 0.7479 & 0.9251 & 0.9111 & 0.7561 & 0.9159 & 0.9161 & 0.7420 \\
    $\blacklozenge$ FAST-VQA \cite{FAST-VQA} 
    & 0.8616 & 0.8330 & 0.6650 & 0.8349 & 0.8381 & 0.6535 & 0.9269 & 0.9353 & 0.7686 \\
    $\blacklozenge$ simpleVQA \cite{simpleVQA}
    & 0.8718 & 0.8743 & 0.6715 & 0.8838 & 0.8911 & 0.6988 & 0.8648 & 0.8906 & 0.6766 \\
    $\blacklozenge$ VSFA \cite{VSFA} 
    & 0.9055 & 0.8895 & 0.7287 & 0.9266 & 0.9099 & 0.7614 & 0.8925 & 0.8883 & 0.7095 \\

    \midrule
    \textbf{3DGSV-Assessor (NR)}
    & \blue{0.9617} & \blue{0.9628} & \blue{0.8367} 
    & \blue{0.9604} & \blue{0.9628} & \blue{0.8316} 
    & \blue{0.9590} & \blue{0.9615} & \blue{0.8211}  \\
    \rowcolor{gray!20}  \textbf{3DGSV-Assessor (RR)}
    & \red{0.9714} & \red{0.9726} & \red{0.8529} & \red{0.9691} & \red{0.9727} & \red{0.8461} & \red{0.9663} & \red{0.9703} & \red{0.8385}  \\

    \bottomrule
  \end{tabular}
}
\label{tab:benchmark}
\vspace{-2mm}
\end{table*}

\subsection{Experiment Setup}
\label{sec:exp_setup}
\subsubsection{\textbf{ Evaluation Datasets and Protocol}}
To quantify the correlation between each method's predicted scores and the corresponding ground-truth MOSs, we adopt three evaluation criteria: Spearman Rank Correlation Coefficient (SRCC), Pearson Linear Correlation Coefficient (PLCC), and Kendall Rank Correlation Coefficient (KRCC). Higher SRCC/PLCC/KRCC indicate better performance. In addition to our own dataset, 3DGS-VBench+, three external NVS VQA benchmarks are incorporated for cross-dataset evaluation: GSC-QA~\cite{GSC-QA}, NeRF-VSQA~\cite{NeRF-VSQA}, and ENeRF-QA~\cite{ENeRF-QA}.

\subsubsection{\textbf{ Reference Algorithms}}
To assess the performance of our proposed method, we select representative VQA metrics for comparison, which can be classified into three groups: (1) Traditional QA Metrics, including: PSNR, SSIM~\cite{PSNR}, LPIPS~\cite{LPIPS}, DISTS~\cite{DISTS}, VIF~\cite{VIF}, FSIM~\cite{FSIM}, IW-SSIM~\cite{IW-SSIM}, MS-SSIM~\cite{MS-SSIM}, and BRISQUE~\cite{BRISQUE}. (2) LMM-based Models, including: LLaVA-one-vision (7B)~\cite{Llava-one-vision}, mPLUG-Owl3 (7B)~\cite{mPLUG-Owl3}, InternVL3.5 (38B/8B)~\cite{InternVL3.5}, InternVL3 (8B)~\cite{InternVL3}, GLM-4.1V (9B)~\cite{GLM-4.1V}, Qwen3-VL (32B/8B)~\cite{Qwen3-VL}, and Qwen3.5 (35B/9B)~\cite{Qwen3.5}. (3) Deep Learning-based VQA Models, including: DOVER~\cite{DOVER}, FAST-VQA~\cite{FAST-VQA}, simpleVQA~\cite{simpleVQA}, and VSFA~\cite{VSFA}.

\subsubsection{\textbf{Training Settings}}
Traditional metrics require no training and are applied directly to the dataset. For FR metrics, each 3DGS model is rendered at the held-out test viewpoints, each rendering is scored against the corresponding ground-truth view, and the scores are averaged into a single quality value per sequence. Since aligned ground-truth videos are generally unavailable along the rendered trajectories, FR scores are computed at these test viewpoints as a proxy.
LMM-based models are first assessed in a zero-shot regime, using each model's publicly released pre-trained weights for direct inference without any dataset-specific fine-tuning. To ensure a rigorous and fair comparison, we further fine-tune two representative LMMs, Llama3.2-Vision (11B)~\cite{llama3.2} and Qwen3.5 (9B)~\cite{Qwen3.5}, under the identical training/testing split on 3DGS-VBench+, allowing for a direct comparison of their adaptability and performance potential. Consistent with their native generation-based formulation, these fine-tuned LMMs are supervised directly on textual score outputs via a cross-entropy objective. 
Deep learning-based VQA models are retrained following their original protocols, using the same dataset split as our framework.
To examine whether our unified, all-in-one framework can match or exceed models specialized for a single dimension, training and inference for these fine-tuned baselines are conducted separately for each of the three quality dimensions.

Our proposed model is implemented in PyTorch and trained on a single 80\,GB NVIDIA A800 GPU, with Qwen3.5 (9B)~\cite{Qwen3.5} serving as the backbone in \texttt{bfloat16}. 
The learning rate is set to $2{\times}10^{-4}$ for the LoRA adapters and $1{\times}10^{-3}$ for the regression/fusion components, both decayed with a cosine annealing schedule. We adopt the AdamW optimizer ($\beta_{1}{=}0.9$, $\beta_{2}{=}0.999$) throughout training.
Stage~1 uses a cross-entropy weight of $\alpha{=}1.0$ and a regression weight linearly warmed to $\beta{=}0.2$ over the first $500$ steps.
Stage~1 and Stage~3 each train for $3$ epochs with an effective batch size of $16$, while Stage~2 trains for up to $10$ epochs with early stopping (patience $3$) using MOS-stratified mini-batches of $12$ renderings drawn from a single scene, so that every batch supplies same-scene ranking pairs. The ranking loss uses a margin of $\rho{=}0.1$ and ignores pairs whose overall-MOS gap falls below $\tau{=}5$. Stage~3 bounds the modulation factor $\boldsymbol{\gamma}\in[1-\epsilon,1+\epsilon]$ with $\epsilon{=}0.3$.
For RR assessment, each scene is paired with a single captured view randomly drawn from the views used to train its 3DGS model (Sec.~\ref{Camera Trajectory Construction for Video Rendering}), fixed once per scene with a fixed random seed and shared by all videos of that scene. The same procedure is applied to the three external benchmarks. The robustness of the results to this random choice is examined in the supplementary material.
Unless otherwise stated, we adopt a within-scene split: for each scene, the rendered videos are partitioned $4{:}1$ into training and testing sets, such that no video appears in both, and results are averaged over 5-fold cross-validation. The three external benchmarks are used only for zero-shot testing, with no fine-tuning.

\subsection{Comparison with Existing Metrics on 3DGS-VBench+}
As reported in Table~\ref{tab:benchmark}, traditional metrics such as PSNR and SSIM correlate only moderately with human judgments and cannot attribute degradation to its geometric or chromatic source, with NR metrics such as BRISQUE performing markedly worse, underscoring the difficulty of assessing 3DGS-specific distortions from low-level statistics alone. 
Larger zero-shot LMM variants generally achieve higher correlations, although the gains vary across models and quality dimensions.
However, LMMs remain insufficient, as their strength in high-level semantic understanding does not translate to sensitivity for low-level compression artifacts. 
Fine-tuning two representative LMMs with the same LoRA-based procedure as ours yields substantial gains, confirming LoRA adaptation as an effective alignment mechanism, while the further margin achieved by our full model, which pairs the same backbone with dedicated regression heads under an additional $\mathcal{L}_reg$ objective, indicates that explicit continuous-valued regression adds perceptual sensitivity beyond LoRA adaptation alone. 
Deep learning-based VQA models such as DOVER and FAST-VQA are competitive but are trained and deployed as separate single-dimension predictors, unable to predict all three dimensional scores jointly.
Both variants of 3DGSV-Assessor outperform every compared method across all nine metrics, with the RR variant leading overall and the NR variant close behind, confirming that the framework remains highly competitive even without access to reference views. Against the strongest baseline on each dimension, 3DGSV-Assessor (RR) improves SRCC by $6.1\%$, $4.6\%$, and $4.3\%$ on three dimensions respectively, confirming that the all-in-one design unifies accurate multi-dimensional assessment while remaining robust across both RR and NR modes, underscoring its adaptability to real-world 3DGS evaluation settings.

\begin{table*}[t]
  \centering
  \vspace{-3mm}
    \caption{Zero-shot cross-dataset correlation performance on multiple benchmarks. Best results are marked in {\red{RED}} and second-best in {\blue{BLUE}}. } 
    \vspace{-2mm}
  \label{tab:method}%
 
  \renewcommand\arraystretch{0.9}
  \resizebox{\textwidth}{!}{
    \begin{tabular}{l|ccc|ccc|ccc}
    \toprule
    \multirow{2}[1]{*}{Method} & \multicolumn{3}{c|}{GSC-QA\cite{GSC-QA}} & \multicolumn{3}{c|}{NeRF-VSQA\cite{NeRF-VSQA}} & \multicolumn{3}{c}{ENeRF-QA\cite{ENeRF-QA}} \\
\cmidrule{2-10}    & SRCC &PLCC & KRCC  & SRCC &PLCC & KRCC  & SRCC &PLCC & KRCC   \\
    \midrule
    DOVER \cite{DOVER} 
    & 0.8174 & 0.7994 & 0.6257
    & 0.6757 & 0.7281 & 0.5146
    & 0.9138 & 0.8991 & 0.7525 \\
    FAST-VQA \cite{FAST-VQA}  
    & 0.6066 & 0.5741 & 0.4427
    & 0.4175 & 0.4922 & 0.3019
    & 0.7842 & 0.7796  & 0.5996 \\
    simpleVQA \cite{simpleVQA} 
    & 0.7307 & 0.7196 & 0.5339
    & 0.2071 & 0.1770 & 0.1520
    & 0.7281 & 0.7611 & 0.5390 \\
    VSFA \cite{VSFA}
    & 0.7159 & 0.7101 & 0.5429
    & 0.5517 & 0.5707 & 0.4043
    & 0.9007 & 0.8862 & 0.7288 \\
    \midrule
    \rowcolor{mycolor_gray} \textbf{3DGSV-Assessor (NR)} 
    & \blue{0.9130} & \blue{0.8950} & \blue{0.7539} 
    & \blue{0.7437} & \blue{0.7796} & \blue{0.5647} 
    & \blue{0.9471} & \blue{0.9296} & \blue{0.8249} \\
    \rowcolor{mycolor_gray} \textbf{3DGSV-Assessor (RR)} 
    & \red{0.9333} &\red{0.9141} & \red{0.7840}
    & \red{0.8180} & \red{0.8036}&\red{0.6452} 
    & \red{0.9644} &\red{0.9393} & \red{0.8401} \\
    \bottomrule
    \end{tabular}%
  }
\label{cross-dataset}
\end{table*}%

\begin{table*}[t]
\vspace{-1mm}
\renewcommand\arraystretch{1}
  \caption{Ablation study on the macro components and LoRA adaptation strategy of 3DGSV-Assessor.}
\vspace{-2mm}
  \resizebox{1\textwidth}{!}{
  \begin{tabular}{ccccc|ccc:ccc:ccc}
    \Xhline{1px}
& \multicolumn{4}{c}{\textbf{Component \& Strategy}}  & \multicolumn{3}{c}{\textbf{Overall Quality}} & \multicolumn{3}{c}{\textbf{Geometry Quality}} &  \multicolumn{3}{c}{\textbf{Color Quality}} \\
    \cline{2-14}
    \multicolumn{1}{c}{No.} & LMM & Ref-Cue & LoRA(vis) & LoRA(llm) & SRCC & PLCC & KRCC & SRCC & PLCC & KRCC & SRCC & PLCC & KRCC \\
    \hline
    \multicolumn{1}{c}{(1)} &  & \ding{52} &  & 
    & 0.8951 & 0.9029 & 0.7171 & 0.8391 & 0.8339 & 0.6448 & 0.8643 & 0.8666 & 0.6764 \\
    \multicolumn{1}{c}{(2)} & \ding{52} & \ding{52} &  &
    & 0.9008 & 0.9012 & 0.7195 & 0.8300 & 0.8448 & 0.6382 & 0.8885 & 0.8864 & 0.6985 \\
    \multicolumn{1}{c}{(3)} & \ding{52} &  & \ding{52} & \ding{52}
    & 0.9617 & 0.9628 & 0.8367 & 0.9604 & 0.9628 & 0.8316 & 0.9590 & 0.9615 & 0.8211 \\
    \multicolumn{1}{c}{(4)} & \ding{52} &  & \ding{52} &
    & 0.9489 & 0.9519 & 0.8216 
    & 0.9502 & 0.9571 & 0.8235 
    & 0.9451 & 0.9491 & 0.8076 \\
    \multicolumn{1}{c}{(5)} & \ding{52} &  &  & \ding{52}
    & 0.9554 & 0.9552 & 0.8251 
    & 0.9472 & 0.9496 & 0.8195 
    & 0.9478 & 0.9523 & 0.8107 \\
    \rowcolor{gray!20} \multicolumn{1}{c}{(6)} & \ding{52} & \ding{52} & \ding{52} & \ding{52}
    & \textbf{0.9714} & \textbf{0.9726} & \textbf{0.8529} & \textbf{0.9691} & \textbf{0.9727} & \textbf{0.8461} & \textbf{0.9663} & \textbf{0.9703} & \textbf{0.8385} \\
    \Xhline{1px}
  \end{tabular}\label{tab:abl_macro}
   }
   \vspace{-2.5mm}
\end{table*}

\subsection{Zero-shot Cross-dataset Evaluation}
\label{sec:exp_cross}

In-domain correlation alone cannot show whether the learned quality scores transfer beyond a single benchmark. We therefore train all learning-based baselines and 3DGSV-Assessor on 3DGS-VBench+ and evaluate them on GSC-QA~\cite{GSC-QA}, NeRF-VSQA~\cite{NeRF-VSQA}, and ENeRF-QA~\cite{ENeRF-QA} without any fine-tuning on these benchmarks.

As shown in Table~\ref{cross-dataset}, both variants of 3DGSV-Assessor outperform every baseline on all three benchmarks, with the RR variant ranking first and the NR variant second on all nine columns.
The baselines are less stable across benchmarks: simpleVQA, competitive on 3DGS-VBench+, drops to an SRCC of $0.2071$ on NeRF-VSQA, and VSFA, the second-strongest baseline on the overall quality of 3DGS-VBench+, drops to $0.5517$ there. DOVER is the most stable baseline but remains behind both variants of our framework on all three benchmarks: relative to DOVER, the NR variant improves SRCC by $11.7\%$, $10.1\%$, and $3.6\%$ on GSC-QA, NeRF-VSQA, and ENeRF-QA respectively, while the RR variant improves SRCC by $14.2\%$, $21.1\%$, and $5.5\%$.

The relative improvement of RR over NR is larger where NR is less accurate: it is $1.0\%$, $1.8\%$, $2.2\%$, and $10.0\%$ on 3DGS-VBench+, ENeRF-QA, GSC-QA, and NeRF-VSQA, where the SRCC of NR is $0.9617$, $0.9471$, $0.9130$, and $0.7437$, respectively. This suggests that the reference cue is more useful when the no-reference prediction is less accurate.

\subsection{Ablation Study}
\subsubsection{\textbf{Effectiveness of components and training strategy}}
We conduct ablation experiments to validate the effectiveness of the macro components and the LoRA adaptation strategy in 3DGSV-Assessor, with results summarized in Table~\ref{tab:abl_macro}. Our analysis reveals three key findings. First, experiments (1), (2), and (6) show that the LMM contributes only once its representation is made quality-aware: experiment (1), regressing the three scores from the reference cue alone, already reaches $0.8951$ SRCC on overall quality, confirming that reference-video agreement is by itself a usable signal, while attaching a frozen LMM (2) yields only a marginal gain and even a small loss on geometry, indicating that a pre-trained backbone's strength in high-level semantic understanding does not by itself transfer to low-level distortion sensitivity. Adapting the backbone with LoRA (6) then improves substantially over both, by $8.5\%$, $15.5\%$, $11.8\%$ and $7.8\%$, $16.8\%$, $8.8\%$ on the three dimensions respectively, confirming that reading quality from the hidden states requires a backbone tuned to distortion rather than to semantics, and that the LMM regressor delivers a substantial accuracy gain over the reference-cue-only pathway once properly adapted.
Second, experiments (3)-(5) validate our joint LoRA strategy: adapting only the vision tower (4) or only the language model (5) each falls short of adapting both together (3), confirming that dual-tower adaptation is necessary for the strongest backbone representation. 
Third, comparing (3) and (6) isolates the reference-aware cue, which improves SRCC by $0.8$--$1.0\%$ across the three dimensions. This gain is modest because the no-reference model is already strong on this in-domain benchmark, and it is larger in the zero-shot cross-dataset evaluation (Sec.~\ref{sec:exp_cross}).
Experiment (6), integrating all three components, achieves the best performance on all nine measures, confirming the necessity of the unified framework for multi-dimensional 3DGS video quality assessment.

\begin{table*}[t]
\centering
\vspace{-3mm}
\caption{Ablation on the number of reference views $K$ used to form the scene anchor $z_R$ (A1$\ldots$AK). Stage-2 columns report cue-level quality: Intra-RankAcc / SRCC$_{\mathrm{cue}}$ are computed \emph{within} each scene ($a_k=\cos(z_V,z_R)$ vs.\ overall MOS; SRCC$_{\mathrm{cue}}$ is the unweighted mean of per-scene Spearman). Global-RankAcc / Global-SRCC$_{\mathrm{cue}}$ ignore scene identity (one pooled Spearman). Overall / Geometry / Color columns report end-to-end quality prediction after gated fusion of the corresponding cue. The best results are marked in \red{RED}.}
\vspace{-2mm}
\label{tab:abl_num_refs}
\setlength{\tabcolsep}{3.0pt}
\renewcommand{\arraystretch}{1.05}
\resizebox{\textwidth}{!}{%
\begin{tabular}{c cccc|cc|cc|cc}
\toprule
\multirow{2}{*}{$K$} & \multicolumn{4}{c|}{\textbf{Stage~2 (cue)}} & \multicolumn{2}{c|}{\textbf{Overall}} & \multicolumn{2}{c|}{\textbf{Geometry}} & \multicolumn{2}{c}{\textbf{Color}} \\
\cline{2-11}
 & Intra-RankAcc$\uparrow$ & SRCC$_{\mathrm{cue}}\uparrow$ & Global-RankAcc$\uparrow$ & Global-SRCC$_{\mathrm{cue}}\uparrow$ & SRCC & PLCC & SRCC & PLCC & SRCC & PLCC \\
\midrule
\rowcolor{gray!15} \textbf{K=1 (ours)} 
& \red{\textbf{0.9683}} & 0.9295 & \red{\textbf{0.9200}} & \red{\textbf{0.8895}}
& \red{\textbf{0.9714}} & \red{\textbf{0.9726}} 
& \red{\textbf{0.9691}} & \red{\textbf{0.9727}} 
& \red{\textbf{0.9663}} & \red{\textbf{0.9703}} \\
K=2 
& 0.9619 & 0.9247 & 0.8616 & 0.7879 
& 0.9702 & 0.9720 
& 0.9678 & 0.9719 
& 0.9655 & 0.9697 \\
K=3 
& 0.9641 & \red{0.9362} & 0.8159 & 0.7008 
& 0.9701 & 0.9718 
& 0.9676 & 0.9717 
& 0.9655 & 0.9695 \\
K=4 
& 0.9565 & 0.9259 & 0.9037 & 0.8679 
& 0.9700 & 0.9717 
& 0.9676 & 0.9716 
& 0.9654 & 0.9696 \\
\bottomrule
\end{tabular}%
}
\end{table*}

\begin{table*}[t]
\centering
\vspace{-2mm}

\caption{Ablation of the Stage-2 training objective. Intra-scene ranking forms ranking pairs only within a scene, whereas global ranking forms them across all scenes in a batch; cross-NCE is a contrastive term that treats same-scene embeddings as positives and other-scene embeddings as negatives. Stage-2 columns report cue-level quality using the cosine agreement $a_i = \cos(\mathbf{z}_{V^{(i)}}, \mathbf{z}_R)$ against overall MOS. RankAcc is the fraction of video pairs with an overall-MOS gap of at least $\tau$ whose cue ordering matches the MOS ordering. Intra-RankAcc and SRCC$_{\text{cue}}$ are computed within each scene and averaged over scenes; Global-RankAcc and Global-SRCC$_{\text{cue}}$ pool all scenes. Stage-3 columns report end-to-end prediction after gated fusion of the corresponding cue. Best results are in \red{red}, second-best in \blue{blue}.}

\vspace{-2mm}
\label{tab:abl_stage2_obj}
\setlength{\tabcolsep}{3.0pt}
\renewcommand{\arraystretch}{1.05}
\resizebox{\textwidth}{!}{%
\begin{tabular}{lcccc|cc|cc|cc}
\toprule
\multirow{2}{*}{\textbf{Variant}} & \multicolumn{4}{c|}{\textbf{Stage~2 (cue)}} & \multicolumn{2}{c|}{\textbf{Overall}} & \multicolumn{2}{c|}{\textbf{Geometry}} & \multicolumn{2}{c}{\textbf{Color}} \\
\cline{2-11}
 & Intra-RankAcc$\uparrow$ & SRCC$_{\mathrm{cue}}\uparrow$ & Global-RankAcc$\uparrow$ & Global-SRCC$_{\mathrm{cue}}\uparrow$
 & SRCC & PLCC & SRCC & PLCC & SRCC & PLCC \\
\midrule
\rowcolor{gray!15}
\textbf{Intra-scene ranking (ours)}
  & \red{\textbf{0.9683}} & \blue{\textbf{0.9295}} & \blue{\textbf{0.9200}} & \blue{\textbf{0.8895}}
  & \red{\textbf{0.9714}} & \red{\textbf{0.9726}}
  & \red{\textbf{0.9691}} & \red{\textbf{0.9727}}
  & \red{\textbf{0.9663}} & \red{\textbf{0.9703}} \\
Global ranking
  & \blue{0.9625} & \red{\textbf{0.9299}} & \red{\textbf{0.9432}} & \red{\textbf{0.9213}}
  & \blue{0.9689} & \blue{0.9710}
  & \blue{0.9689} & \blue{0.9722}
  & \blue{0.9651} & \blue{0.9690} \\
Intra-scene ranking + cross-NCE
  & 0.9087 & 0.8445 & 0.6772 & 0.4211
  & 0.9686 & 0.9708
  & 0.9680 & 0.9724
  & 0.9649 & 0.9682 \\
Global ranking + cross-NCE
  & 0.9504 & 0.9165 & 0.8890 & 0.8514
  & 0.9676 & 0.9697
  & 0.9674 & 0.9720
  & 0.9656 & 0.9685 \\
\bottomrule
\end{tabular}%
}
\vspace{-2mm}
\end{table*}

\begin{table*}[t]
\centering
\renewcommand\arraystretch{1.05}
\caption{Ablation on the Stage-3 cue-fusion mechanism. Stage-1 (NR) is the no-reference baseline that predicts from $\mathbf{u}$ alone; the other variants inject the reference cue $\mathbf{c}$ through different fusion operators. \textbf{Heads} indicates whether the Stage-1 regression heads are kept frozen or retrained after fusion. The best results are marked in \red{RED}.}
\vspace{-2mm}
\label{tab:abl_stage3_full}
\resizebox{\textwidth}{!}{%
\begin{tabular}{ll c|ccc:ccc:ccc}
\toprule
\multirow{2}{*}{\textbf{Variant}} & \multirow{2}{*}{\textbf{Fusion} $\tilde{\mathbf{u}}$} & \multirow{2}{*}{\textbf{Heads}} & \multicolumn{3}{c}{\textbf{Overall}} & \multicolumn{3}{c}{\textbf{Geometry}} & \multicolumn{3}{c}{\textbf{Color}} \\
\cmidrule(lr){4-6}\cmidrule(lr){7-9}\cmidrule(lr){10-12}
 & & & SRCC & PLCC & KRCC & SRCC & PLCC & KRCC & SRCC & PLCC & KRCC \\
\midrule
      Stage-1 (NR)
      & $\mathbf{u}$ & frozen
      & 0.9617 & 0.9628 & 0.8367 
      & 0.9604 & 0.9628 & 0.8316 
      & 0.9590 & 0.9615 & 0.8211\\
    Concatenate cue
      & $[\mathbf{u}\Vert\mathcal{A}(\mathbf{c})]$ & retrained
      & 0.9618 & 0.9669 & 0.8306
      & 0.9666 & 0.9709 & 0.8389
      & 0.9584 & 0.9673 & 0.8210 \\
    Sigmoid gate
      & $\mathbf{u}\odot\sigma(\mathcal{A}(\mathbf{c}))$ & frozen
      & 0.9688 & 0.9702 & 0.8469
      & 0.9680 & 0.9711 & 0.8430
      & 0.9641 & 0.9694 & 0.8347 \\
\midrule
    \rowcolor{gray!20} \textbf{Bounded gate (ours)}
      & \textbf{$\mathbf{u}\odot\bigl(\mathbf{1}+\epsilon\tanh(\mathcal{A}(\mathbf{c}))\bigr)$} & \textbf{frozen}
      & \red{\textbf{0.9714}} & \red{\textbf{0.9726}} & \red{\textbf{0.8529}}
      & \red{\textbf{0.9691}} & \red{\textbf{0.9727}} & \red{\textbf{0.8461}}
      & \red{\textbf{0.9663}} & \red{\textbf{0.9703}} & \red{\textbf{0.8385}} \\
\bottomrule
\end{tabular}%
}
\end{table*}

\subsubsection{\textbf{Effect of the Number of Reference Views}}
\label{app:abl_num_refs}

Table~\ref{tab:abl_num_refs} varies the number of reference views $K$ aggregated into the scene anchor $\mathbf{z}_R$, where $K{=}1$ is the default single-reference encoder and $K{>}1$ averages the embeddings of $K$ training viewpoints.
Two observations support the single-view design. First, at the cue level, $K{=}1$ attains the best intra-scene ranking, with an Intra-RankAcc of 0.9683, and the best cross-scene consistency, with a Global-RankAcc of 0.9200 and a Global-SRCC$_{\mathrm{cue}}$ of 0.8895, whereas multi-view anchors lower the global metrics. 
The only metric where a multi-view anchor is ahead is the per-scene SRCC$_{\mathrm{cue}}$ at $K{=}3$ (0.9362 against 0.9295), a gap that does not carry over to any downstream measure. Second, at the system level, Stage-3 prediction is nearly insensitive to $K$, with an SRCC spread below 0.002 on every dimension and $K{=}1$ best throughout.
Since $K{=}1$ is also the cheapest setting, requiring only a single view per scene to encode, store, and transmit in the RR mode, we adopt it as the default.

\subsubsection{\textbf{Effect of the Stage-2 Training Objective}}
\label{app:abl_stage2_obj}

Table~\ref{tab:abl_stage2_obj} compares four Stage-2 training objectives: our default intra-scene ranking, a global ranking variant that enforces MOS-consistent ordering across scenes, and each of the two combined with a cross-scene InfoNCE~\cite{InfoNCE} term.

The central result is that intra-scene ranking is the best objective for the system as a whole: it achieves the highest Stage-3 performance on every reported prediction metric, together with the best Intra-RankAcc of 0.9683. 
Global ranking, as expected, attains the best scene-agnostic cue consistency, with a Global-RankAcc of 0.9432 and a Global-SRCC$_{\mathrm{cue}}$ of 0.9213, yet this advantage does not transfer downstream. A possible reason is that Stage~3 pairs each video only with its own scene's anchor, so it does not directly rely on the cross-scene ordering of the cue. Meanwhile, global ranking slightly lowers Intra-RankAcc (0.9625 against 0.9683) and brings no downstream gain, so the simpler intra-scene objective is preferred.
Adding cross-scene InfoNCE is harmful at the cue level: on top of intra-scene ranking, it lowers Global-SRCC$_{\mathrm{cue}}$ from 0.8895 to 0.4211 and Intra-RankAcc to 0.9087. A possible reason is that the InfoNCE term separates embeddings by scene identity, which may compete with the quality-ordering objective. Nevertheless, Stage-3 performance varies little across the four objectives, with an Overall SRCC between 0.9676 and 0.9714, all above the no-reference baseline of 0.9617. This is consistent with the bounded gate (Sec.~\ref{sec:stage3}), which limits how far a weakened cue can perturb the Stage-1 representation. Taken together, these results support training Stage~2 with the intra-scene ranking objective alone.

\subsubsection{\textbf{Effect of the Stage-3 Cue-Fusion Mechanism}}
\label{app:abl_stage3_fusion}

Table~\ref{tab:abl_stage3_full} compares our bounded gate with the cue-free Stage-1 (NR) baseline and two alternative ways of injecting the reference cue: feature concatenation and a plain sigmoid gate.
Two contrasts are informative. First, the injection form matters. Concatenation feeds $\mathcal{A}(\mathbf{c})$ to retrained heads as an extra input, but its gains over the NR baseline are mixed: PLCC improves on all three dimensions, whereas Overall SRCC is nearly unchanged, Color SRCC falls slightly below the baseline (0.9584 against 0.9590), and Overall KRCC decreases.
In contrast, both multiplicative gates improve all nine metrics while keeping the Stage-1 heads frozen. As these variants differ in both the fusion operator and whether the heads are retrained, this suggests that modulating the existing quality representation can be a more reliable way to use the cue.
Second, constraining the modulation further improves performance. The sigmoid gate scales $\mathbf{u}$ by a factor in $(0,1)$, which can only attenuate the representation the frozen heads were calibrated on and never reproduces it exactly; our gate instead applies an identity-centered, bounded factor $\boldsymbol{\gamma}\in[1-\epsilon,1+\epsilon]$, which reduces to the Stage-1 representation when $\mathcal{A}(\mathbf{c})$ outputs zero. This variant is the best on all nine metrics, consistent with both design choices: multiplicative modulation over concatenation, and an identity-centered bounded gate over a purely attenuating one.

%% file: sec/6_conclusion.tex
\section{Conclusion}
In this paper, we introduce 3DGS-VBench+, a large-scale, multi-dimensional subjective VQA benchmark for compressed 3DGS, comprising 3{,}036 3DGS reconstructions, each paired with a rendered video, across 11 real-world and 22 synthetic scenes, six representative compression algorithms, and 9{,}108 MOSs over the overall, geometry, and color dimensions. 
Built on it, we propose 3DGSV-Assessor, an LMM-based framework that progresses from a no-reference base to reference-flexible assessment: an LMM-based NR regressor that jointly predicts all three quality dimensions. A reference-aware ranking encoder learns a quality-related cue from the rendered video and a reference view supervised by human scores, and a bounded gate makes the reference optional, refining the scores when a view is available and reverting exactly to the no-reference prediction otherwise. Experiments show SOTA correlation with human perception on 3DGS-VBench+ under both settings and strong zero-shot generalization to other NVS benchmarks. We hope this work serves as a testbed and a diagnostic tool for developing perceptually grounded evaluation and optimization of 3DGS.